\documentclass{article}

\usepackage{microtype}
\usepackage{graphicx}
\usepackage{subcaption}
\usepackage{booktabs} 

\usepackage{hyperref}

\usepackage[accepted]{icml2026}
\usepackage{amsmath}
\usepackage{amssymb}
\usepackage{mathtools}
\usepackage{amsthm}
\usepackage{booktabs}
\usepackage{multirow}
\usepackage{graphicx}
\usepackage[most]{tcolorbox}
\definecolor{bg_gray}{RGB}{245,245,245}
\definecolor{border_gray}{RGB}{200,200,200}

\usepackage[capitalize,noabbrev]{cleveref}

\theoremstyle{plain}

\theoremstyle{definition}

\theoremstyle{remark}

\usepackage[disable,textsize=tiny]{todonotes}

\begin{document}

\twocolumn[
  \icmltitle{Endogenous Reprompting: Self-Evolving \\
   Cognitive Alignment for Unified Multimodal Models
}



  \icmlsetsymbol{equal}{*}
  \begin{icmlauthorlist}
    \icmlauthor{Zhenchen Tang}{yyy,comp}
    \icmlauthor{Songlin Yang}{sch}
    \icmlauthor{Zichuan Wang}{yyy,comp}
    \icmlauthor{Bo Peng}{yyy,equal}
    \icmlauthor{Yang Li}{yyy,comp}
    \icmlauthor{Beibei Dong}{yyy,comp}
    \icmlauthor{Jing Dong}{yyy,equal}

  \end{icmlauthorlist}

  \icmlaffiliation{yyy}{Department of XXX, University of YYY, Location, Country}
  \icmlaffiliation{comp}{Company Name, Location, Country}
  \icmlaffiliation{sch}{School of ZZZ, Institute of WWW, Location, Country}

  \icmlcorrespondingauthor{Firstname1 Lastname1}{first1.last1@xxx.edu}
  \icmlcorrespondingauthor{Firstname2 Lastname2}{first2.last2@www.uk}

  \icmlkeywords{Machine Learning, ICML}

\begin{center}
\small
$^{1}$ New Laboratory of Pattern Recognition, Institute of Automation, Chinese Academy of Sciences \\
$^{2}$ School of Artificial Intelligence, University of Chinese Academy of Sciences \\
$^{3}$ MMLab@HKUST, The Hong Kong University of Science and Technology
\end{center}


  \vskip 0.3in
]



\printAffiliationsAndNotice{}  

\begin{abstract}
Unified Multimodal Models (UMMs) exhibit strong understanding, yet this capability often fails to effectively guide generation. We identify this as a Cognitive Gap: the model lacks the understanding of how to enhance its own generation process. To bridge this gap, we propose Endogenous Reprompting, a mechanism that transforms the model’s understanding from a passive encoding process into an explicit generative reasoning step by generating self-aligned descriptors during generation. To achieve this, we introduce SEER (Self-Evolving Evaluator and Reprompter), a training framework that establishes a two-stage endogenous loop using only 300 samples from a compact proxy task, Visual Instruction Elaboration. First, Reinforcement Learning with Verifiable Rewards (RLVR) activates the model’s latent evaluation ability via curriculum learning, producing a high-fidelity endogenous reward signal. Second, Reinforcement Learning with Model-rewarded Thinking (RLMT) leverages this signal to optimize the generative reasoning policy. Experiments show that SEER consistently outperforms state-of-the-art baselines in evaluation accuracy, reprompting efficiency, and generation quality, without sacrificing general multimodal capabilities. Code and project page are available at https://2kxx.github.io/SEER.github.io/.

\end{abstract}

\section{Introduction}

\begin{figure}[t]
  \centering
    \includegraphics[width=0.45\textwidth]{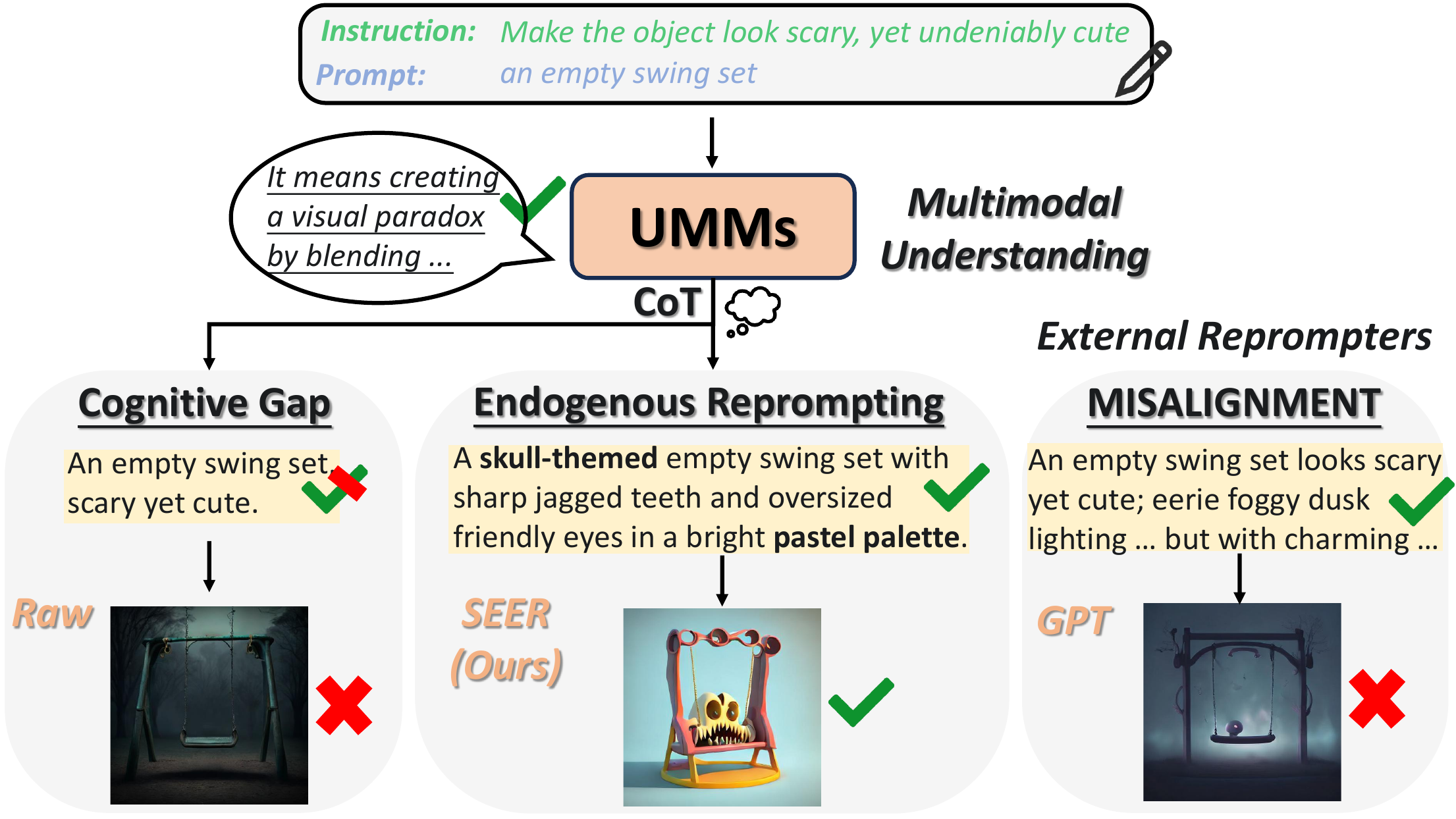}
    \caption{Bridging the cognitive gap. While direct generation fails to reflect the model's understanding (Left), and external reprompters often cause misalignment by generating descriptors that mismatch the generator's priors (Right), SEER (Middle) leverages Endogenous Reprompting. It produces concrete, self-aligned descriptors that strictly match the generator's generative priors, successfully bridging understanding and generation.}
    \label{fig1}
\vspace{-0.6cm}
\end{figure}

Recent Unified Multimodal Models (UMMs) have successfully integrated understanding and generation capabilities into a single framework \cite{xie2024show, dong2023dreamllm}. By sharing a common backbone, UMMs possess the potential for seamless cross-modal interaction, offering a promising path to resolving the semantic disconnect often found in traditional separate-encoder T2I models. However, a significant imbalance persists: despite a shared architecture, the model's ability to generate lags substantially behind its ability to understand \cite{chen2025janus, pan2025transfer, xie2025show, wang2024emu3, jin2025srum}. As illustrated in Figure \ref{fig1}, while the model accurately comprehends visual instructions, it struggles to translate this understanding into generative guidance, resulting in misalignment in the final output. We identify this as a cognitive gap: the model lacks the specific understanding of how to enhance its own generation process (details in the Appendix \ref{app:failure_analysis}).

To bridge this gap, we propose Endogenous Reprompting, a mechanism that transforms the model’s understanding from a passive encoding process into an explicit generative reasoning step by translating user intention into self-aligned descriptors (aligned with the model's own capabilities) during generation. Prior approaches \cite{betker2023improving, wang2025promptenhancer} rely on disjoint models (e.g., an LLM and a generator), which leads to representation mismatch, causing the generated descriptions to mismatch the generator’s priors in Figure \ref{fig1}. In contrast, our approach exploits the intrinsic representational consistency of UMMs. Being endogenous, all reasoning operates within the model’s shared representation space, enabling knowledge to be seamlessly transferred between understanding and generation to generate reprompts. This ensures Model-Specific Alignment: the model translates user intention into specific linguistic descriptors that are aligned with the generator's generative priors. We argue that this constitutes the missing ``understanding-to-generation'' pathway.

To achieve this, we propose SEER (Self-Evolving Evaluator and Reprompter), a training framework that establishes a two-stage endogenous loop using only 300 samples from a compact proxy task, Visual Instruction Elaboration. We introduce this task to specifically bootstrap generative reasoning: it targets the explicit elaboration of implicit user intention, forcing the model to reason about visual instructions to enhance generation rather than simply mapping text to pixels. Using this compact signal, SEER unfolds in two stages. First, we employ Reinforcement Learning with Verifiable Rewards (RLVR) \cite{shao2024deepseekmath, wen2025reinforcement} through a curriculum learning strategy to activate the internal evaluator. By adopting a pairwise optimization strategy that learns from relative comparisons, we transform the model's understanding into a high-fidelity reward signal.
Second, we employ Reinforcement Learning with Model-rewarded Thinking (RLMT) \cite{bhaskar2025language} to harness this signal, enabling the model to autonomously refine its reasoning policy. This distinguishes our work from traditional Reinforcement learning from human feedback (RLHF) for visual generation \cite{ziegler2019fine, xu2023imagereward, liu2025flow, gong2025onereward}: whereas RLHF targets the generation execution phase (optimizing pixels), our approach optimizes the generative reasoning process (optimizing prompts), effectively teaching the model to think before it generates.

Empirically, SEER enhances instruction compliance and reasoning capabilities, utilizing the model's own understanding to resolve the cognitive gap. The entire self-evolution process relies on minimal data, highlighting the efficiency of unlocking intrinsic potential rather than learning from scratch. Our key contributions are summarized as follows:
\begin{itemize}
\item We pioneer Endogenous Reprompting, establishing the first generative reasoning framework for UMMs to bridge the cognitive gap via a compact proxy task, Visual Instruction Elaboration.
    
\item We propose SEER, a self-evolving framework that establishes a two-stage endogenous loop to achieve Endogenous Reprompting, activating an internal evaluator with RLVR and optimizing the reasoning policy with RLMT.
    
\item We demonstrate that SEER consistently outperforms state-of-the-art baselines in instruction compliance, reprompting efficiency, and generation quality, using only 300 samples, without sacrificing general multimodal capabilities.
\end{itemize}

\begin{figure*}[t]
    \centering
    \includegraphics[width=0.9\textwidth]{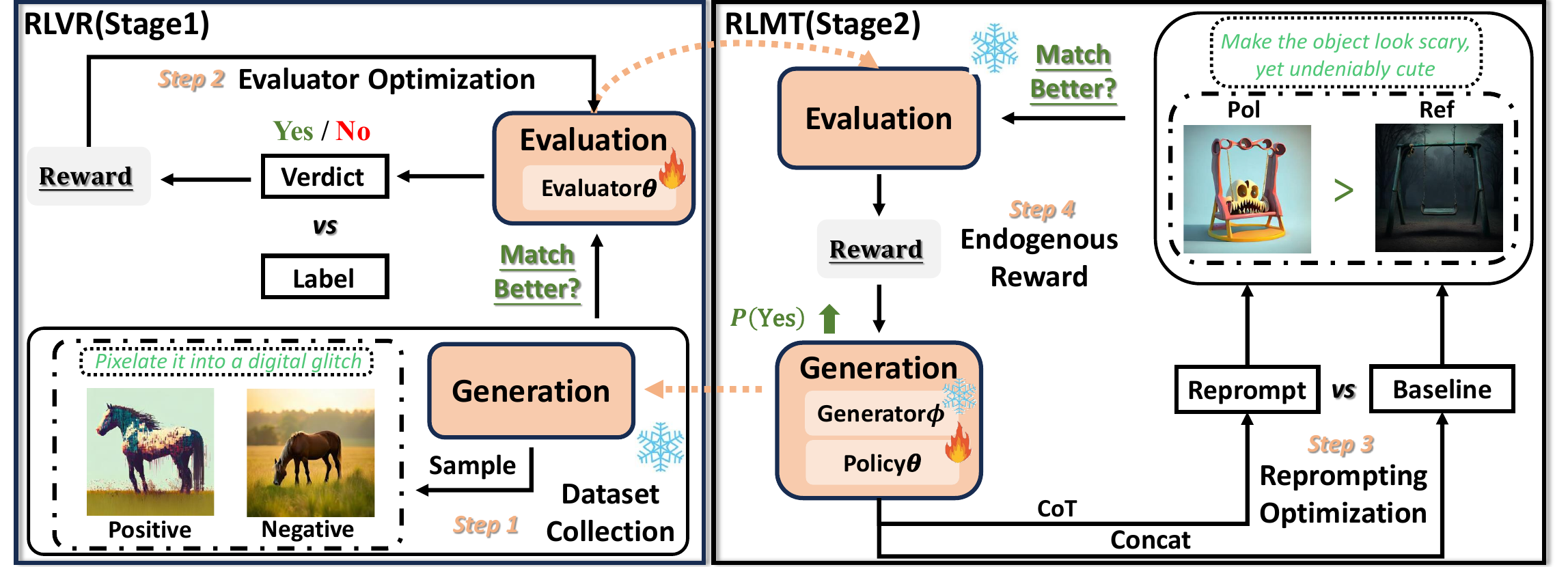}
    \caption{
Overview of SEER. The framework bridges the cognitive gap via a two-stage endogenous loop. RLVR: (Step 1) collecting endogenous image pairs to (Step 2) optimize the internal evaluator. RLMT: (Step 3) optimizing the Reprompting Policy using (Step 4) relative rewards derived from the activated evaluator. 
    }
    \label{fig2}
\vspace{-0.6cm}
\end{figure*}

\section{Related Work}
\noindent \textbf{Unified Multimodal Models.} 
UMMs integrate understanding and generation within a single end-to-end framework to promote cross-modal interaction. Recent research explores a broad architectural design space, primarily categorized into three paradigms: (1) Autoregressive Models, such as Janus \cite{wu2025janus}, and Emu3 \cite{wang2024emu3}, which tokenize inputs for sequential next-token prediction, with variants like Show-o \cite{xie2024show} incorporating discrete diffusion schedules; (2) Hybrid Architectures, which either attach diffusion heads to a shared transformer backbone \cite{zhao2024monoformer, chen2024diffusion} or route features from frozen MLLMs to external generators via learnable queries \cite{chen2025blip3}; and (3) Emerging Designs, including Masked-Autoregressive (MAR) generation \cite{li2024autoregressive, wu2025harmonizing} and scalable Mixture-of-Transformers (MoT) frameworks \cite{zhou2024transfusion, deng2025emerging}. Despite these architectural strides, a common bottleneck persists: the alignment between strong understanding and generation remains insufficiently activated, creating the cognitive gap that SEER aims to bridge.

\noindent \textbf{Prompt Optimization and Reprompting.} 
To bridge the gap between user intent and generative execution, researchers have increasingly prioritized Prompt Optimization. Foundational approaches like DALL-E 3 \cite{betker2023improving} utilize recaptioning to generate dense captions for model fine-tuning. For inference-time enhancement, on-the-fly rewriting has become a dominant paradigm, employing LLMs for iterative critique and refinement \cite{wu2024self}, automated prompt engineering \cite{cao2023beautifulprompt}, or complex agent-based planning \cite{wang2024genartist}. Recent works like PromptEnhancer \cite{wang2025promptenhancer} further advance this by training dedicated rewriters against fine-grained reward models. However, these methods typically couple disjoint models, leading to a fundamental representation mismatch. In contrast, by leveraging the intrinsic representational consistency of UMMs, SEER activates an endogenous mechanism to generate reprompts aligned with the generative priors (i.e., model-specific alignment).

\noindent \textbf{Reinforcement Learning for Visual Generation.} 
Aligning generative models with human preferences is a critical research frontier. Early works like Imagereward \cite{xu2023imagereward} fine-tune diffusion models using preference scores, while Diffusion-DPO \cite{wallace2024diffusion} directly optimizes policies without explicit reward models. Further advances like DDPO \cite{su2024ddpo} and GRPO \cite{liu2025flow} formulate denoising as a decision-making problem, with recent works like OneReward \cite{gong2025onereward} and RewardDance \cite{wu2025rewarddance} leveraging VLMs to provide feedback. Crucially, existing methods primarily focus on the execution phase by optimizing diffusion trajectories. SEER diverges by applying RLMT to the generative reasoning phase. We demonstrate that evolving the generative reasoning (reprompt) rather than the low-level execution offers a more effective pathway to unlock the generative potential of UMMs .

\section{Method}
In this section, we introduce SEER (Self-Evolving Evaluator and Reprompter), a framework designed to bridge the cognitive gap in UMMs (Figure \ref{fig2}). Our goal is to transform the model's multimodal understanding into an active generative reasoning capability via a self-evolving loop.

\subsection{Problem Formulation}

We formalize the task of Visual Instruction Elaboration as a reasoning optimization problem. Let a visual instruction be denoted by $a \in \mathcal{A}$ (e.g., ``Make the object look scary, yet undeniably cute'') and an initial minimal prompt by $p_0 \in \mathcal{P}$ (e.g., ``an empty swing set''). Our objective is to leverage the model's latent capabilities to generate a Reprompt $p \in \mathcal{P}$. Derived from an explicit reasoning path, this reprompt bridges the cognitive gap by translating the user intention of $a$ into concrete descriptors that are executable by the generator. The generation process yields an image $x_{\text{pol}} = G(p)$ that satisfies the instruction $a$ while maintaining the semantic integrity of $p_0$.

We conceptualize the UMM as comprising understanding/reasoning parameters $\theta$ and generation parameters $\phi$. To verify that the improvement stems from optimizing the ``understanding-to-generation'' pathway rather than enhancing low-level rendering capabilities, we freeze the generation parameters $\phi$ and exclusively optimize $\theta$. The model operates in three endogenous functional modes:
\begin{itemize}
    \item \textbf{Generator} $G(\cdot | \phi): \mathcal{P} \to \mathcal{X}$, which maps textual prompts to the image space (Fixed).
    \item \textbf{Evaluator} $E(\cdot | \theta): \mathcal{X}^2 \times \mathcal{T} \to [Yes, No]$, which acts as a pairwise judge that estimates the preference probability $P(x_{\text{pol}} \succ x_{\text{ref}} \mid \text{context})$. Here, $\mathcal{T}$ is the textual condition space. $x_{\text{ref}} = G([p_0; a])$ is the baseline.
    \item \textbf{Reprompting Policy} $\pi_\theta(\cdot | \theta): \mathcal{P} \times \mathcal{A} \to \Delta(\mathcal{P})$, which generates the reprompt via reasoning.
\end{itemize}

Formally, we define the optimization reward $R$ as the relative preference probability against a naive baseline. For an image pair $\mathbf{x} = (x_{\text{pol}}, x_{\text{ref}})$, the reward integrates three intrinsic dimensions into a holistic judgment:
\begin{equation}
\label{eq:reward_components}
R(p; a, p_0) \triangleq \underbrace{E(\mathbf{x}, a)}_{\text{Compliance}} + \underbrace{E(\mathbf{x}, p_0)}_{\text{Consistency}} + \underbrace{E(\mathbf{x})}_{\text{Quality}},
\end{equation}
Here, $E(\mathbf{x}, a)$ quantifies adherence to the visual instruction, $E(\mathbf{x}, p_0)$ ensures fidelity to the original prompt, and $E(\mathbf{x})$ assesses general aesthetic quality. 

To internalize this reasoning capability, we optimize the policy parameters $\theta$ to maximize the expected reward while preventing deviation from the original language distribution:
\begin{equation}
\label{eq:objective}
\theta^* = \arg\max_\theta \mathbb{E}_{p \sim \pi_\theta(\cdot \mid a, p_0)} \Big[ R(p; a, p_0) \Big] - \lambda \, D_{\mathrm{KL}}(\pi_\theta \| \pi_{\mathrm{ref}}).
\end{equation}
where $\pi_{\mathrm{ref}}$ represents the reference policy (the initial model), and $D_{\mathrm{KL}}$ acts as a regularization term to ensure the fluency of the generated reprompts.

\begin{figure*}[t]
    \centering
    \begin{minipage}{0.48\textwidth}
        \centering
        \includegraphics[width=\textwidth]{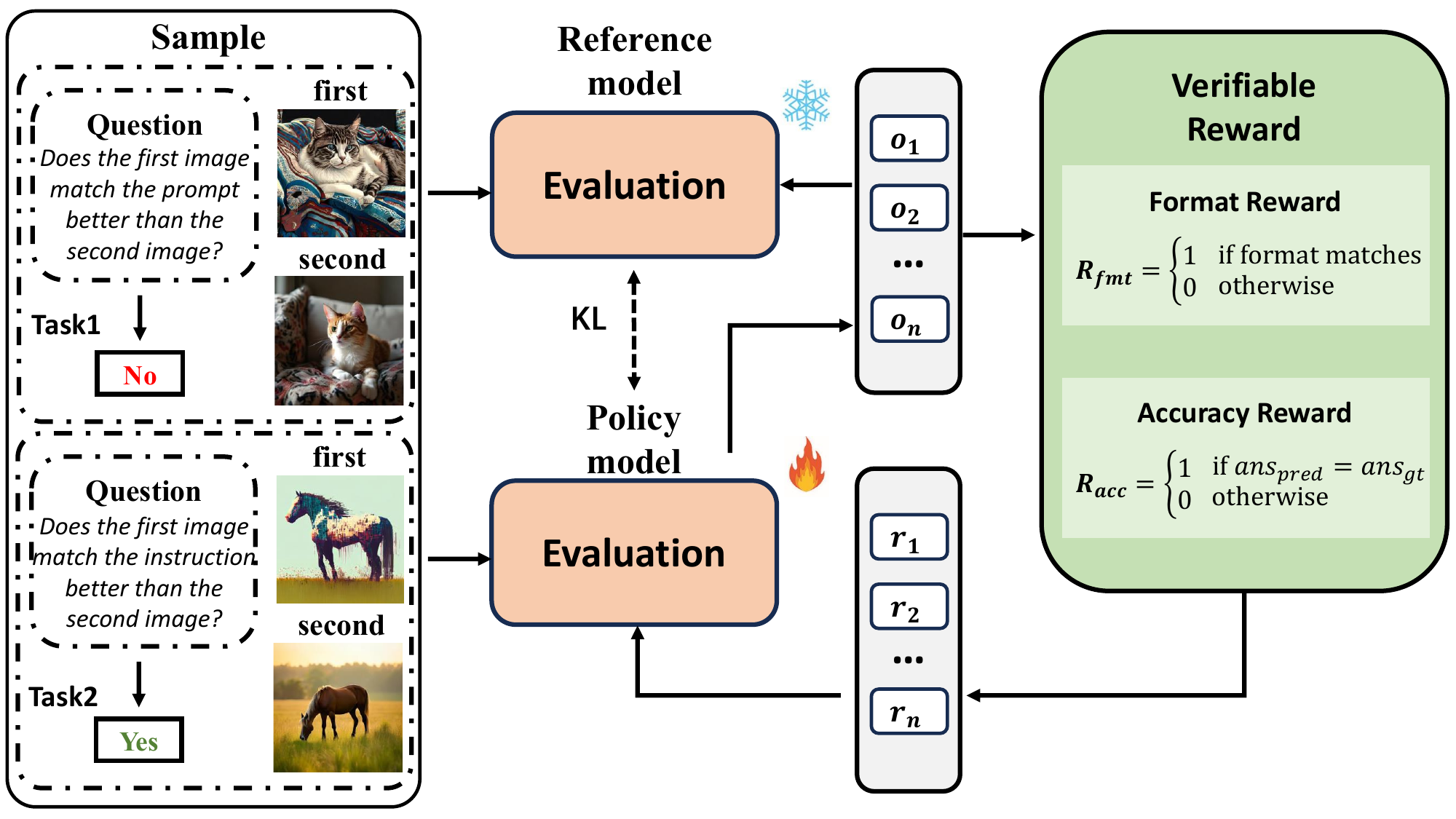}
        \caption{Stage 1: RLVR. We employ curriculum learning to transform the model into a robust internal critic. By training on pairwise comparisons (from basic alignment to instruction discrimination) using GRPO, we activate a high-fidelity internal evaluator $E(x; a, p_0)$ capable of assessing user intention.}
        \label{fig3}
    \end{minipage}
    \hfill
    \begin{minipage}{0.48\textwidth}
        \centering
        \includegraphics[width=\textwidth]{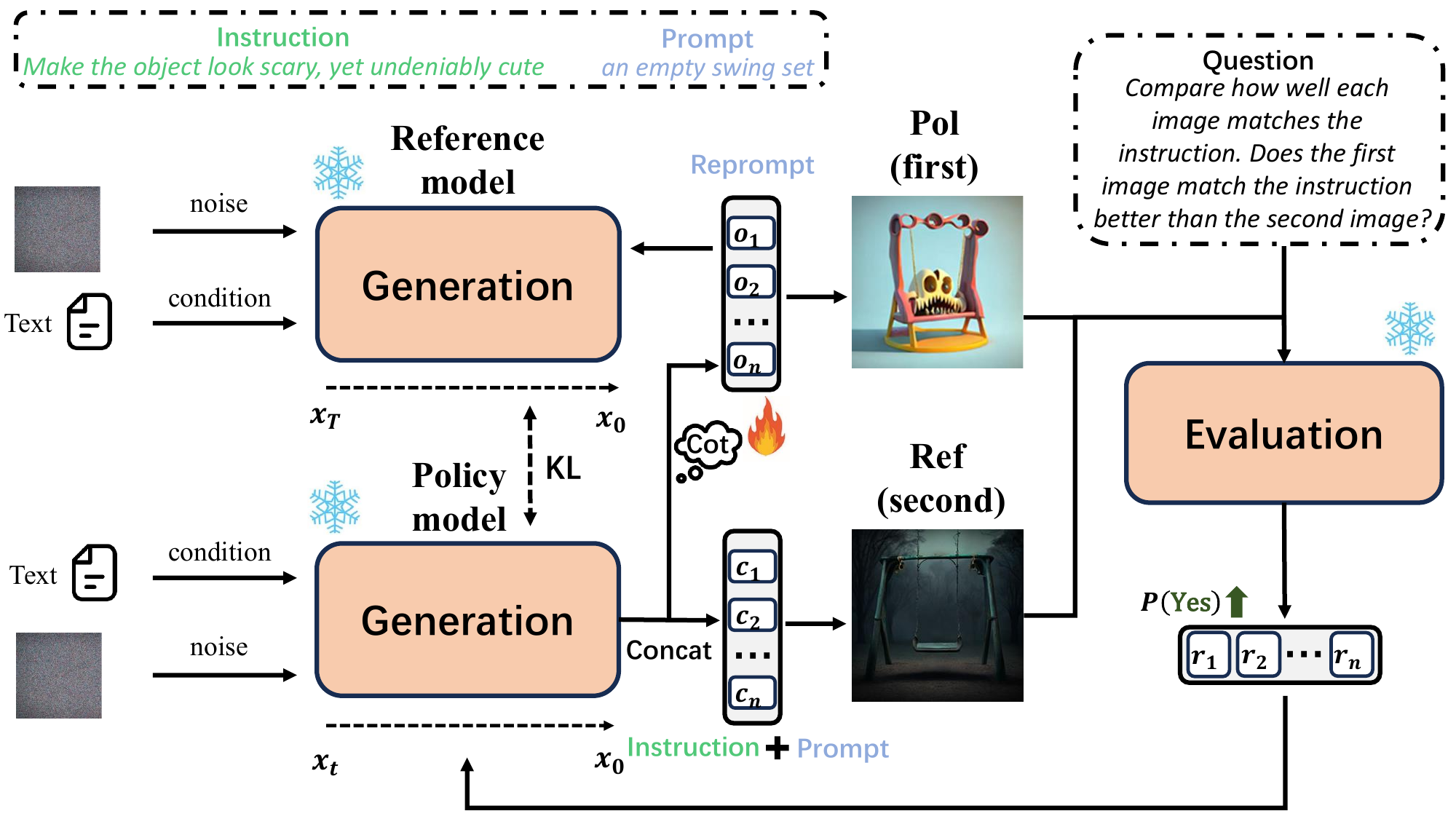}
        \caption{Stage 2: RLMT. The model optimizes its reasoning policy $\pi_\theta$ via a ``Think-Generate-Evaluate'' loop. The endogenous reward is computed by comparing the reasoned generation ($x_{\text{pol}}$) against a naive baseline ($x_{\text{ref}}$), effectively teaching the model to ``think'' (reprompt) for superior visual results.}
        \label{fig4}
    \end{minipage}
    \vspace{-0.3cm}
\end{figure*}

\subsection{Preliminary}
Reinforcement Learning from Human Feedback (RLHF) \cite{ziegler2019fine} aims to align model outputs with human preferences. Given an input $i \sim \mathcal{I}$, the model generates a response $o \sim \pi_\theta(\cdot \mid i)$. A reward function $R$ trained on human preferences assigns a score to each response: $r(o, i) \in \mathbb{R}$. RLHF optimizes $\theta$ to maximize expectations:
\begin{equation}
\max_\theta \; \mathbb{E}_{i \sim \mathcal{I}} \; \mathbb{E}_{o \sim \pi_\theta(\cdot \mid i)} \big[ R(i, o) \big].
\end{equation}
Reinforcement Learning with Verifiable Reward (RLVR) \cite{wen2025reinforcement} is a paradigm for training models when verification is possible. It modifies the RLHF framework by replacing the model-based reward $r$ with a verification function, e.g., the indicator $\mathbf{1}\{o = y\}$ against a ground-truth answer $y$. Meanwhile, models in RLVR often generate an intermediate reasoning trace $z \sim \pi_\theta(\cdot \mid i)$ followed by a final response $o \sim \pi_\theta(\cdot \mid i, z)$. The training objective maximizes the expected correctness of the final response:
\begin{equation}
\max_\theta \; \mathbb{E}_{i\sim\mathcal{I}} \; \mathbb{E}_{(o,z)\sim \pi_\theta(\cdot\mid i)} \big[ \mathbf{1}\{o = y\} \big].
\end{equation}

Reinforcement Learning with Model-rewarded Thinking (RLMT) \cite{bhaskar2025language} fuses the reasoning-oriented structure of RLVR with the reward function of RLHF. It requires the model to generate a reasoning trace $z$ before producing the final response $o$, while utilizing a learned reward model $R$ for scoring instead of rule-based verification. The objective is to maximize the expected reward:
\begin{equation} 
    \max_\theta \; \mathbb{E}_{i\sim\mathcal{I}} \; \mathbb{E}_{(o,z)\sim \pi_\theta(\cdot\mid i)} \big[ R(i, o) \big]. 
\end{equation}

\subsection{Visual Instruction Elaboration}
\label{AIG}
To bootstrap generative reasoning efficiently, we construct a compact dataset $\mathcal{D} = \{(a^{(i)}, p_0^{(i)})\}_{i=1}^{300}$. We define this as a proxy task: its purpose is not to cover all possible instructions, but to provide high-leverage samples that bootstrap the model's generative reasoning.
We categorize these samples based on the reasoning depth required to bridge the cognitive gap. Simple Instructions focus on direct attribute mapping, covering Material Synthesis, Perspective Shifts, and Semantic Edits (e.g., explicit material synthesis like ``carved of ice''). Hard Instructions necessitate deep reasoning to interpret conceptual demands, covering Attribute Reasoning, Constraint Reasoning, and Conceptual Reasoning (e.g., ``Make the object look scary, yet undeniably cute''). The details are in Appendix \ref{app:dataset}.

The dataset is instantiated into two formats to align with our two-stage training framework. For RLVR, we expand the inputs into quadruplets $\mathcal{D}_{\text{RLVR}} = \{(a_i, p_{0,i}, \mathbf{x}_i, y_i)\}_{i=1}^{300}$. Here, $\mathbf{x}_i = (x^+_i, x^-_i)$ denotes a pair of candidate images collected with minimal human supervision, and $y_i \in \{\textsc{Yes}, \textsc{No}\}$ serves as a binary validity label. For RLMT, we utilize the raw inputs $\mathcal{D}_{\text{RLMT}} = \mathcal{D}$ directly, requiring the model to autonomously generate a reasoning chain.

\subsection{Self-Evolving Training Framework}
\label{sec:framework}

We propose SEER, a self-evolving training framework for efficient and automatic endogenous optimization. Instead of relying on external supervision, the framework forms a two-stage loop: RLVR activates the model’s latent understanding into a verifiable reward, and RLMT uses this endogenous signal to evolve the reprompting policy.

\subsubsection{Stage 1: Reinforcement Learning with Verifiable Rewards (RLVR)}

The goal of this stage is to transform the model’s multimodal understanding into a high-fidelity endogenous reward signal. We leverage the model to consolidate the three reward dimensions defined in Eq.~\ref{eq:reward_components} into a holistic evaluation function $E(x; a, p_0)$ (in Appendix \ref{reward_prompt}), employing Reinforcement Learning with Verifiable Rewards (RLVR) (Figure \ref{fig3}).

\noindent \textbf{Curriculum Learning.}

Directly training on visual instructions can lead to unstable gradients. We therefore implement a curriculum to progressively ``wake up'' the evaluator:

Stage 1: Basic alignment. 
We train on concrete image--text pairs. The evaluator learns to judge basic visual--semantic alignment, establishing a foundational decision boundary.

Stage 2: Visual Instruction Supervision.
We train on quadruplets from $\mathcal{D}_{\text{RLVR}}$. The evaluator must identify which image better satisfies the visual instruction $a$ while preserving $p_0$, bridging the cognitive gap verifiably.

\noindent \textbf{Pairwise Evaluator Optimization.}

We formulate the evaluator's output generation as a sequential decision process and adopt pairwise formulation to mitigate reward hacking \cite{wang2025pref}. Given a query $q = (a, p_0)$ and a candidate image pair $\mathbf{x}$, the policy $\pi_\theta$ generates a reasoning trace $z_i$ followed by a response (verdict) $o_i$.
We optimize $\pi_\theta$ using Group Relative Policy Optimization (GRPO) \cite{shao2024deepseekmath} over the entire trajectory $u_i = [z_i; o_i] = (u_{i,1}, u_{i,2}, \dots, u_{i,L_i})$. For each query, we sample $N$ trajectories $\{u_i\}_{i=1}^N$ from $\pi_{\theta_{\text{old}}}$. The objective maximizes the advantage-weighted probability per token:
\begin{equation}
\label{eq:token_grpo}
\begin{aligned}
&\mathcal{J}_{\text{GRPO}}(\theta) = \mathbb{E}_{q, \mathbf{x}, \{u_i\}} \Bigg[ \frac{1}{N} \sum_{i=1}^N \frac{1}{L_i} \sum_{t=1}^{L_i} \Bigg( \min(s_{i,t}^1 A_i, s_{i,t}^2 A_i) \\
&\quad - \beta D_{\mathrm{KL}} \left( \pi_\theta(u_{i,t} | u_{i,<t}, q, \mathbf{x}) \| \pi_{\mathrm{ref}}(u_{i,t} | u_{i,<t}, q, \mathbf{x}) \right) \Bigg) \Bigg],
\end{aligned}
\end{equation}
where $u_{i,t}$ denotes the $t$-th token of the $i$-th trajectory. The token-level importance sampling ratios are defined as:
\begin{equation}
s_{i,t}^1 = \frac{\pi_\theta(u_{i,t} \mid u_{i,<t}, q, \mathbf{x})}{\pi_{\theta_{\text{old}}}(u_{i,t} \mid u_{i,<t}, q, \mathbf{x})}, \quad
s_{i,t}^2 = \mathrm{clip}\Big(s_{i,t}^1, 1-\epsilon, 1+\epsilon\Big).
\end{equation}
The advantage $A_i$ is computed at the sequence level by normalizing the scalar rewards within the group:
\begin{equation}
A_i = \frac{r_i - \mathrm{mean}(\{r_j\}_{j=1}^N)}{\mathrm{std}(\{r_j\}_{j=1}^N)}.
\end{equation}
The scalar reward $r_i$ is derived from the binary verification result: $r_i = \mathbb{I}(\text{valid format}) + \mathbb{I}(o_i = y)$.

\subsubsection{Stage II: Reinforcement Learning with Model-rewarded Thinking (RLMT)}

The goal of this stage is to enhance the model's generative reasoning policy. To achieve this, we employ Reinforcement Learning with Model-rewarded Thinking (RLMT) and leverages the RLVR-tuned evaluator as a high-fidelity endogenous reward model (Figure \ref{fig4}).

\noindent \textbf{Endogenous Relative Reward.}
To derive a robust scalar reward $r_i$ for each sampled reprompt $o_i$, we establish a comparative baseline using a naive prompt $c_i$. This baseline is constructed by directly concatenating the visual instruction $a$ and the initial prompt $p_0$ (denoted as $c_i = [p_0; a]$), representing a stable and competitive bound. We argue that a successful reprompt must yield visual results superior to this naive concatenation. Accordingly, for a given input $(a, p_0)$, we generate two distinct images:

Reasoned Generation: $x_{\text{pol}} = G(o_i)$, produced using the generated reprompt $o_i$.

Naive Baseline: $x_{\text{ref}} = G(c_i)$, produced using the direct concatenation $c_i$.

The total reward $r_i$ aggregates visual preference and format validity. We define the accuracy reward $R_{\text{acc}}$ by leveraging the RLVR-tuned evaluator $E$ to compare the images. Since the evaluator is trained to output a binary verdict, the reward is defined as an indicator function:
\begin{equation}
    R_{\text{acc}} = \mathbb{I}\left( E(x_{\text{pol}}, x_{\text{ref}}; a, p_0) = \textsc{Yes} \right).
\end{equation}
This assigns a reward of 1 if the reasoned generation $x_{\text{pol}}$ is judged superior to the baseline $x_{\text{ref}}$, and 0 otherwise. The final reward is $r_i = R_{\text{acc}} + \mathbb{I}(\text{valid format})$. This mechanism drives the policy to discover reprompts that offer tangible visual improvements over trivial prompt engineering.

\noindent \textbf{Reprompting Policy Optimization.}
We optimize the reasoning policy $\pi_\theta$ using the identical GRPO objective (Eq.~\ref{eq:token_grpo}). The primary adaptation lies in the input dependency: here, the model generates the reprompt sequence $o_i$ conditioned solely on the textual query $q = (a, p_0)$, without visual context $\mathbf{x}$. Consequently, the probability terms in both the importance sampling ratio and the KL divergence simplify to depend only on the textual context. The token-level ratio becomes:
\begin{equation}
    s_{i,t}^1 = \frac{\pi_\theta(u_{i,t} \mid u_{i,<t}, q)}{\pi_{\theta_{\text{old}}}(u_{i,t} \mid u_{i,<t}, q)}.
\end{equation}
The regularization becomes $D_{\mathrm{KL}} ( \pi_\theta(\cdot|q) \| \pi_{\mathrm{ref}}(\cdot|q) )$.
Substituting the total reward $r_i$ into the advantage calculation aligns the gradient updates with Eq.~\ref{eq:objective}:
\begin{equation}
    \theta^* = \arg\max_\theta \mathbb{E}_{p \sim \pi_\theta(\cdot \mid a, p_0)} \Big[ r_i \Big] - \lambda \, D_{\mathrm{KL}}(\pi_\theta \| \pi_{\mathrm{ref}}).
\label{rlmt}
\end{equation}
Crucially, since $r_i$ relies on the contrast between the reasoned output and the naive baseline, this objective biases the policy towards non-trivial reasoning. It forces the model to discover reprompts that yield visual outcomes strictly superior to those of the direct concatenation. 

This strategy marks a fundamental shift from traditional RLHF applied to visual generation. Whereas standard RLHF targets the generation execution phase (optimizing pixels), SEER targets the generative reasoning process (optimizing prompts). By optimizing the prompts rather than the pixels, SEER effectively teaches the model to ``think'' before it generates, thereby internalizing the reasoning process and bridging the cognitive gap.

\subsubsection{Mechanistic Analysis: Implicit Model-Specific Alignment}
\label{sec:analysis}

A critical advantage of SEER lies in its structural enforcement of model-specific alignment. Let $\mathcal{Z}_{\text{gen}}$ denote the latent content space of generative priors that the generator $G$ can reliably realize.
By combining disjoint models, prior methods suffer from a fundamental representation mismatch: the prompter generates descriptions without knowing if the generator can actually visualize them, and even when tuned with an external reward $R_{\text{sem}}$, they optimize a signal that fails to accurately assess the generator's capabilities:
\begin{equation}
    p^*_{\text{ext}} = \arg\max_p R_{\text{proxy}}(G(p), a).
\end{equation}
Consequently, the generated reprompts $p^*_{\text{ext}}$ often drift away from $\mathcal{Z}_{\text{gen}}$, leading to descriptions that are linguistically valid but visually unrealizable. 
In contrast, Endogenous Reprompting resolves this mismatch leveraging the intrinsic representational consistency of UMMs. While Eq.~\ref{eq:objective} explicitly optimizes a relative preference, we formalize this as a proxy for likelihood maximization under the Bradley-Terry assumption \cite{bradley1952rank}. By modeling the preference probability as $P(x_{\text{pol}} \succ x_{\text{ref}}) = \sigma(r(x_{\text{pol}}) - r(x_{\text{ref}}))$ where the reward $r(x)$ approximates alignment $\log P_{\theta}(a|x)$, the objective can be interpreted as:
\begin{equation}
    \mathcal{J}_{\text{SEER}} \approx \mathbb{E}_{p \sim \pi_\theta} \left[ \log P_{\theta}(a \mid G(p; \phi)) - \beta \log \frac{\pi_\theta(p)}{\pi_{\text{ref}}(p)} \right].
\end{equation}
Critically, the reliability of this proxy stems from the unified architecture. Although the evaluator is only fine-tuned on limited samples, it inherits the massive pre-trained knowledge. Since $E$ and $G$ share the same representation space, the evaluator retains inherent sensitivity to the generator's failure modes. It acts as a validity filter that naturally penalizes prompts lying outside the generator's capabilities. Thus, the reprompting policy learns to navigate the unified representation space, evolving reprompts into self-aligned descriptions that satisfy the generator's priors. 
Formally, for any reprompt $\hat{p} \notin \mathcal{Z}_{\text{gen}}$ that mismatches the generator’s priors, the shared priors enforce a penalty:
\begin{equation}
   G(\hat{p}) \text{ fails} \implies P_{\theta}(a \mid G(\hat{p})) \to 0 \implies \pi_\theta(\hat{p}) \to 0.
\end{equation}
This drives the optimal policy $\pi^*_{\text{SEER}}$ to converge to the strict intersection of linguistic intent and generative feasibility:
\begin{equation}
    \text{supp}(\pi^*_{\text{SEER}}) \subseteq \{ p \mid p \in \mathcal{P} \land G(p) \text{ realizes } a \}.
\end{equation}

\section{Experiments}
\label{sec:experiments}

In this section, we evaluate SEER's ability to bridge the cognitive gap. We first detail the experimental setup, then present the core results on Visual Instruction Elaboration (evaluating both the internal judge and the generative reasoning), and finally demonstrate that our method preserves the model's general multimodal capabilities.

\subsection{Experimental Setup}

\subsubsection{Base Model Architecture.} 
We implement our framework on Harmon, a Unified Multimodal Model (UMM) with 1.5B parameters. Harmon adopts a masked autoregressive (MAR) generation paradigm, utilizing a pre-trained MAE \cite{li2024autoregressive} as the visual understanding encoder and a VAE \cite{kingma2013auto} as the visual generation tokenizer. During training, we exclusively optimize the parameters governing multimodal understanding and reasoning, while keeping the visual generation components frozen. RecA \cite{xie2025reconstruction} demonstrates that Harmon can achieve high-fidelity reconstruction by conditioning solely on its own understanding embeddings. This latent connectivity is crucial: it implies that the ``Internal Evaluator'' and ``Generator'' share a grounded feature space, fulfilling the prerequisite for model-specific alignment derived in Sec.~\ref{sec:analysis}.

\subsubsection{Implementation Details.} 
We conduct all training experiments on 2 NVIDIA A100 (40GB) GPUs using the AdamW \cite{loshchilov2017decoupled} optimizer. 
For Stage 1 (RLVR), we employ a staged learning rate schedule: $1 \times 10^{-6}$ for the Basic Alignment phase and $3 \times 10^{-6}$ for the Visual Instruction phase. 
For Stage 2 (RLMT), the learning rate is set to $1 \times 10^{-6}$. 
In the GRPO objective, we set the number of iterations per update to 1, the KL divergence coefficient $\beta$ to 0.04, and the group size $N=6$. 
Both stages are trained for 10 epochs. The entire self-evolution process is highly data-efficient, utilizing only 300 samples.

\subsubsection{Evaluation Benchmarks.}
To rigorously evaluate our framework, we construct a dedicated Visual Instruction Test set using the same categories defined in Sec.~\ref{AIG}. We further split it into In-Distribution (ID) (same instructions, different prompts) and Out-of-Distribution (OOD) (different instructions and prompts) subsets to assess generalization in Appendix \ref{difference}. 
Additionally, we evaluate general visual understanding capabilities on MME \cite{fu2025mme}, POPE \cite{li2023evaluating}, GQA \cite{hudson2019gqa}, MMMU \cite{yue2024mmmu}, and SEEDBench \cite{li2023seed}, and assess general generation capabilities on GenEval \cite{ghosh2023geneval} and DPG-Bench \cite{hu2024ella}.

\begin{table*}[t] 
\centering
\caption{Evaluator Accuracy on the Visual Instruction Test Set. We report results on both In-Distribution and Out-Of-Distribution splits, further categorized into Simple (Simp) and Hard subsets based on reasoning difficulty.}
\begin{tabular}{lccccccc} 
\toprule
\multirow{2}{*}{\textbf{Model Config}} & \multirow{2}{*}{\textbf{Total}} & \multicolumn{3}{c}{\textbf{In-Distribution}} & \multicolumn{3}{c}{\textbf{Out-Of-Distribution}} \\
\cmidrule(lr){3-5} \cmidrule(lr){6-8}
 &  & \textbf{Overall} & \textbf{Simp} & \textbf{Hard} & \textbf{Overall} & \textbf{Simp} & \textbf{Hard} \\
\midrule
Zero-shot & 0.41 & 0.44 & 0.40 & 0.47 & 0.38 & 0.35 & 0.40 \\
Phase 1 Only & 0.49 & 0.48 & 0.45 & 0.50 & 0.50 & 0.65 & 0.40 \\
\textbf{SEER-Eval (Ours)} & \textbf{0.92} & \textbf{0.96} & \textbf{1.00} & \textbf{0.93} & \textbf{0.88} & \textbf{0.85} & \textbf{0.90} \\
\bottomrule
\end{tabular}
\label{tab:evaluator_detailed}
\vspace{-0.1cm}
\end{table*}

\begin{table*}[t]
\centering
\caption{Win Ratio of SEER vs. External Reprompting Methods. We report SEER's win rate against each baseline across different reasoning depths. SEER dominates on Hard/Complex tasks while maintaining strict efficiency (\# Avg. Words).}
\begin{tabular}{l c ccc ccc c}
\toprule
\multirow{2.5}{*}{\textbf{External Baseline}} & \multirow{2.5}{*}{\textbf{Total}} & \multicolumn{3}{c}{\textbf{In-Distribution}} & \multicolumn{3}{c}{\textbf{Out-Of-Distribution}} & \multirow{2.5}{*}{\textbf{\# Avg. Words}} \\
\cmidrule(lr){3-5} \cmidrule(lr){6-8}
 & & \textbf{Overall} & \textbf{Simp} & \textbf{Hard} & \textbf{Overall} & \textbf{Simp} & \textbf{Hard} & \\
\midrule
vs. SEER-Gen + BeautifulPrompt & 0.90 & 0.89 & 0.89 & 0.89 & 0.90 & 0.93 & 0.88 & 55.13  \\
vs. SEER-Gen + PromptEnhancer & 0.75 & 0.77 & 0.68 & 0.83 & 0.73 & 0.70 & 0.74 & 153.04 \\
vs. SEER-Gen + GPT-5.2 & 0.68 & 0.73 & 0.69 & 0.76 & 0.62 & 0.61 & 0.63 & 46.23  \\
vs. SEER-Gen + Gemini3 & 0.61 & 0.63 & 0.51 & 0.71 & 0.58 & 0.56 & 0.61 & 33.06  \\
vs. SEER-Gen + Qwen3max & 0.65 & 0.63 & 0.57 & 0.68 & 0.68 & 0.66 & 0.70 & 30.76  \\
\midrule
\textbf{SEER (Ours)} & - & - & - & - & - & - & - & \textbf{22.94} \\
\bottomrule
\end{tabular}
\label{tab:external_comparison}
\vspace{-0.4cm}
\end{table*}

\subsection{Main Results: Visual Instruction Elaboration}

\subsubsection{Evaluator Performance (Stage 1)}
We evaluate the internal evaluator on a human-labeled test set of quadruplets.
Baselines: (1) Zero-shot: Original Harmon model; (2) Phase 1 Only: Trained on basic alignment only; (3) SEER-Eval: Full two-phase curriculum.

\noindent \textbf{Results.} Table~\ref{tab:evaluator_detailed} presents the accuracy across In-Distribution and Out-Of-Distribution sets, further split into Simple (Simp) and Hard subsets.
The Zero-shot model performs near random guessing (0.41), confirming that latent understanding does not automaticallly translate to verification capability.
Phase 1 Only improves basic alignment but stalls on instruction tasks (0.49).
In contrast, SEER-Eval achieves a remarkable 0.92 overall accuracy. Notably, it maintains 0.90 on OOD Hard instructions. This confirms that the ``Compact Proxy Task'' effectively bootstraps the latent evaluation capability, enabling the model to generalize to unseen concepts without massive-scale supervision.

\subsubsection{Generative Reasoning Performance (Stage 2)}

We evaluate the final generation quality via blind pairwise human evaluation. 

\noindent \textbf{Metric: Human Win Ratio.} 
To quantify performance, we conduct blind pairwise comparisons. Annotators present with an image from SEER and a baseline, selecting the one that better satisfies the visual instruction $a$ while preserving the semantic subject of $p_0$. A Win Ratio $>0.5$ indicates that SEER is preferred (details in the Appendix \ref{app:human_eval}). 

\noindent \textbf{Comparison with External Reprompters.}
We benchmark SEER against two categories: (1) Specialized Reprompters: BeautifulPrompt \cite{cao2023beautifulprompt}, PromptEnhancer \cite{wang2025promptenhancer}; and (2) SOTA MLLMs: GPT-5.2 \cite{openai_gpt52}, Gemini3 \cite{google_gemini3}, Qwen3max \cite{alibaba_qwen3max}. For a fair comparison, we substitute SEER's reprompting policy with these external models, feeding their generated reprompts directly into SEER's generator.

\begin{figure*}[t]
    \centering
    \includegraphics[width=1\textwidth]{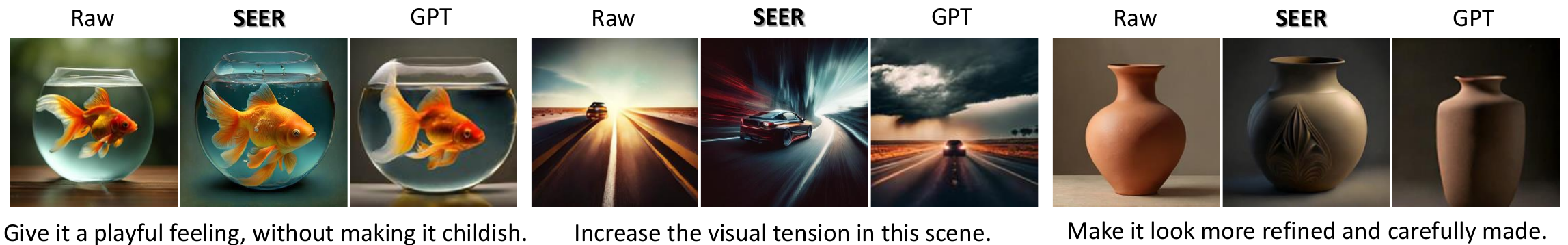}
    \caption{
    Qualitative Comparison. While the Base Model (Left) fails to execute visual instructions due to the cognitive gap, and External Reprompters (Right) cause representation mismatch, SEER (Center) generates self-aligned reprompts that strictly match the generator's priors, yielding superior visual fidelity.}
    \label{fig:qualitative_comparison}
\vspace{-0.2cm}
\end{figure*}

\begin{table*}[t]
\centering
\caption{Human Win Ratio of SEER (Ours) against various baselines. A score $>$ 0.5 indicates SEER is preferred. Results are reported across In-Distribution and Out-Of-Distribution splits, further categorized into Simple (Simp) and Hard subsets.}
\begin{tabular}{l c ccc ccc}
\toprule
\multirow{2.5}{*}{\textbf{Opponent Model}} & \multirow{2.5}{*}{\textbf{Total}} & \multicolumn{3}{c}{\textbf{In-Distribution}} & \multicolumn{3}{c}{\textbf{Out-Of-Distribution}} \\
\cmidrule(lr){3-5} \cmidrule(lr){6-8}
 & & \textbf{Overall} & \textbf{Simp} & \textbf{Hard} & \textbf{Overall} & \textbf{Simp} & \textbf{Hard} \\
\midrule
vs. Base (CoT) & 0.85 & 0.85 & 0.78 & 0.90 & 0.85 & 0.87 & 0.84 \\
vs. Bagel & 0.75 & 0.77 & 0.64 & 0.86 & 0.73 & 0.63 & 0.79 \\
vs. Bagel-Think & 0.69 & 0.77 & 0.63 & 0.86 & 0.62 & 0.55 & 0.66 \\
vs. Blip3-o & 0.81 & 0.80 & 0.76 & 0.83 & 0.83 & 0.93 & 0.76 \\
vs. Show-o2 & 0.81 & 0.85 & 0.78 & 0.89 & 0.77 & 0.73 & 0.80 \\
\bottomrule
\end{tabular}
\label{tab:win_ratio}
\vspace{-0.2cm}
\end{table*}

\begin{table*}[!t]
\centering
\caption{General Generation Capabilities on GenEval and DPG-Bench. Note: For these simple prompts, we disable the reprompting mechanism to test the direct generation quality. SEER maintains or improves fundamental compositional capabilities.}
\resizebox{0.95\linewidth}{!}{
\begin{tabular}{l|ccccccc|c}
\toprule
\multirow{2}{*}{\textbf{Model}} & \multicolumn{7}{c|}{\textbf{GenEval}} & \multirow{2}{*}{\textbf{DPG-Bench}} \\
\cmidrule(lr){2-8}
 & \textbf{Single Obj.} & \textbf{Two Obj.} & \textbf{Counting} & \textbf{Colors} & \textbf{Position} & \textbf{Color Attri.} & \textbf{Overall} & (\textbf{Score}) \\
\midrule
Harmon (Base) & 0.99 & 0.86 & 0.68 & 0.85 & 0.43 & 0.48 & 0.72 & 80.5 \\
Stage 1 Only & 0.99 & 0.87 & 0.68 & 0.86 & 0.44 & 0.48 & 0.72 & 80.9 \\
\textbf{Stage 2 (SEER)} & \textbf{1.00} & \textbf{0.88} & \textbf{0.70} & \textbf{0.87} & \textbf{0.44} & \textbf{0.52} & \textbf{0.74} & \textbf{81.1} \\
\bottomrule
\end{tabular}
}
\label{tab:generation}
\vspace{-0.5cm}
\end{table*}

\noindent \textbf{Results.} As shown in Table~\ref{tab:external_comparison}, SEER consistently outperforms both groups.
It surpasses specialized reprompters (0.90 against BeautifulPrompt), confirming that representation mismatch in disjoint models causes reprompters to mismatch the generator’s priors. 
SOTA MLLMs perform competitively (0.61 against Gemini3), benefiting from stronger understanding knowledge; however, SEER still achieves superior performance with significantly fewer words (22.94 on average).
This efficiency provides empirical support for our mechanistic analysis in Sec.~\ref{sec:analysis}. Notably, SEER's advantage is often amplified on Hard Instructions (e.g., vs PromptEnhancer: 0.74 on Hard vs 0.70 on Simple). As visualized in Figure~\ref{fig:qualitative_comparison}, unlike external LLMs that generate generic, verbose descriptions, SEER identifies descriptors that are aligned with the generator's generative priors. The internal evaluator naturally penalizes misaligned reprompts, pruning the reasoning chain to be maximally executable.

\noindent \textbf{Comparison with UMMs.}
We further compare SEER (1.5B) with state-of-the-art UMMs: Harmon (Base), Bagel (14B) \cite{deng2025emerging}, Bagel-Think (CoT), Blip3-o (8B) \cite{chen2025blip3}, and Show-o2 (7B) \cite{xie2025show}.

\textbf{Results.} Table~\ref{tab:win_ratio} reports the Human Win Ratio of SEER against each baseline.
First, SEER achieves a dominant win ratio (e.g., 0.85) against the Harmon Base, confirming that Endogenous Reprompting effectively bridges the cognitive gap.
Second, despite its smaller parameter size (1.5B), SEER remains competitive against significantly larger models. This competitiveness is largely driven by SEER's robustness on Hard Instructions (e.g., vs Bagel: 0.79 on Hard vs 0.63 on Simple).
Critically, this high alignment with human preferences proves that our Internal Evaluator (Stage 1) successfully evolved to mimic human judgment, thereby correctly guiding the Reprompting Policy (Stage 2) to generate self-aligned reprompts without external supervision.

\subsection{General Capabilities}

\noindent \textbf{General Visual Understanding.} 
To verify that SEER preserves the core cognitive capabilities of UMMs, we evaluate our post-trained model on MME, POPE, GQA, MMMU, and SEEDBench. 
As shown in Table~\ref{tab:understanding}, SEER's performance remains highly stable compared to the base Harmon model, with fluctuations falling within normal fine-tuning variance.
Notably, on MME, SEER achieves 1179 (vs. 1155 base) showing slight improvements.
This stability confirms that RLVR acts as a targeted activation mechanism for the internal evaluator, rather than a destructive overwrite.

\begin{table}[h]
\centering
\caption{General Multimodal Understanding Benchmarks. We report results across progressive training stages. SEER (Stage 2) maintains or slightly improves understanding capabilities compared to the Harmon base model.}
\resizebox{\linewidth}{!}{
\begin{tabular}{lcccccc}
\toprule
\textbf{Model} & \textbf{MME} & \textbf{POPE$_{\text{Acc}}$} & \textbf{POPE$_{\text{F1}}$} & \textbf{GQA} & \textbf{MMMU} & \textbf{SEED} \\
\midrule
Harmon (Base) & 1155 & 83.8 & 83.9 & 58.8 & 34.7 & 65.2 \\
Phase 1 Only & 1172 & 83.7 & 83.9 & 58.9 & 34.8 & 65.2 \\
Stage 1 (RLVR) & \textbf{1179} & \textbf{84.0} & \textbf{84.3} & 58.9 & \textbf{35.2} & 64.4 \\
\textbf{Stage 2 (SEER)} & \textbf{1179} & 83.8 & 84.1 & \textbf{58.9} & 35.1 & \textbf{65.4} \\
\bottomrule
\end{tabular}
}
\label{tab:understanding}
\vspace{-0.3cm}
\end{table}

\noindent \textbf{General Generation.} 
We assess standard text-to-image generation capabilities using GenEval and DPG-Bench. 
Since these benchmarks primarily consist of direct prompts to evaluate basic compositional attributes, we explicitly bypass the reprompting enhancement and evaluate the generator's direct execution.
As shown in Table~\ref{tab:generation}, even without active reasoning, SEER (Stage 2) achieves slight improvements across most metrics (e.g., GenEval Overall 0.72 $\to$ 0.74).
Importantly, these results are achieved without any direct fine-tuning on the image generation objective. 
This confirms that the alignment between understanding and generation is intrinsically refined while learning generative reasoning.

\section{Conclusion}

We address the cognitive gap in UMMs by introducing SEER, a self-evolving framework that transforms passive understanding into active reasoning. Through a two-stage endogenous loop (RLVR and RLMT), SEER bootstraps the model's reprompting capability using only 300 samples via Visual Instruction Elaboration.
Experiments confirm that SEER outperforms baselines in evaluation accuracy, reprompting efficiency, and generation quality while preserving general capabilities. 
Crucially, our approach achieves model-specific alignment, ensuring reasoning acts as precise triggers strictly within the generator's priors. 
This work establishes a new paradigm: shifting focus from optimizing execution to evolving cognitive reasoning.

\bibliography{example_paper}

@article{xie2024show,
  title={Show-o: One single transformer to unify multimodal understanding and generation},
  author={Xie, Jinheng and Mao, Weijia and Bai, Zechen and Zhang, David Junhao and Wang, Weihao and Lin, Kevin Qinghong and Gu, Yuchao and Chen, Zhijie and Yang, Zhenheng and Shou, Mike Zheng},
  journal={arXiv preprint arXiv:2408.12528},
  year={2024}
}

@article{dong2023dreamllm,
  title={Dreamllm: Synergistic multimodal comprehension and creation},
  author={Dong, Runpei and Han, Chunrui and Peng, Yuang and Qi, Zekun and Ge, Zheng and Yang, Jinrong and Zhao, Liang and Sun, Jianjian and Zhou, Hongyu and Wei, Haoran and others},
  journal={arXiv preprint arXiv:2309.11499},
  year={2023}
}

@article{chen2025janus,
  title={Janus-pro: Unified multimodal understanding and generation with data and model scaling},
  author={Chen, Xiaokang and Wu, Zhiyu and Liu, Xingchao and Pan, Zizheng and Liu, Wen and Xie, Zhenda and Yu, Xingkai and Ruan, Chong},
  journal={arXiv preprint arXiv:2501.17811},
  year={2025}
}

@article{pan2025transfer,
  title={Transfer between modalities with metaqueries},
  author={Pan, Xichen and Shukla, Satya Narayan and Singh, Aashu and Zhao, Zhuokai and Mishra, Shlok Kumar and Wang, Jialiang and Xu, Zhiyang and Chen, Jiuhai and Li, Kunpeng and Juefei-Xu, Felix and others},
  journal={arXiv preprint arXiv:2504.06256},
  year={2025}
}

@article{xie2025show,
  title={Show-o2: Improved Native Unified Multimodal Models},
  author={Xie, Jinheng and Yang, Zhenheng and Shou, Mike Zheng},
  journal={arXiv preprint arXiv:2506.15564},
  year={2025}
}

@article{wang2024emu3,
  title={Emu3: Next-token prediction is all you need},
  author={Wang, Xinlong and Zhang, Xiaosong and Luo, Zhengxiong and Sun, Quan and Cui, Yufeng and Wang, Jinsheng and Zhang, Fan and Wang, Yueze and Li, Zhen and Yu, Qiying and others},
  journal={arXiv preprint arXiv:2409.18869},
  year={2024}
}

@article{jin2025srum,
  title={Srum: Fine-grained self-rewarding for unified multimodal models},
  author={Jin, Weiyang and Niu, Yuwei and Liao, Jiaqi and Duan, Chengqi and Li, Aoxue and Gao, Shenghua and Liu, Xihui},
  journal={arXiv preprint arXiv:2510.12784},
  year={2025}
}

@article{betker2023improving,
  title={Improving image generation with better captions},
  author={Betker, James and Goh, Gabriel and Jing, Li and Brooks, Tim and Wang, Jianfeng and Li, Linjie and Ouyang, Long and Zhuang, Juntang and Lee, Joyce and Guo, Yufei and others},
  journal={Computer Science. https://cdn. openai. com/papers/dall-e-3. pdf},
  volume={2},
  number={3},
  pages={8},
  year={2023}
}

@article{wang2024genartist,
  title={Genartist: Multimodal llm as an agent for unified image generation and editing},
  author={Wang, Zhenyu and Li, Aoxue and Li, Zhenguo and Liu, Xihui},
  journal={Advances in Neural Information Processing Systems},
  volume={37},
  pages={128374--128395},
  year={2024}
}

@article{wang2025promptenhancer,
  title={Promptenhancer: A simple approach to enhance text-to-image models via chain-of-thought prompt rewriting},
  author={Wang, Linqing and Xing, Ximing and Cheng, Yiji and Zhao, Zhiyuan and Li, Donghao and Hang, Tiankai and Tao, Jiale and Wang, Qixun and Li, Ruihuang and Chen, Comi and others},
  journal={arXiv preprint arXiv:2509.04545},
  year={2025}
}

@article{shao2024deepseekmath,
  title={Deepseekmath: Pushing the limits of mathematical reasoning in open language models},
  author={Shao, Zhihong and Wang, Peiyi and Zhu, Qihao and Xu, Runxin and Song, Junxiao and Bi, Xiao and Zhang, Haowei and Zhang, Mingchuan and Li, YK and Wu, Yang and others},
  journal={arXiv preprint arXiv:2402.03300},
  year={2024}
}

@article{wen2025reinforcement,
  title={Reinforcement learning with verifiable rewards implicitly incentivizes correct reasoning in base llms},
  author={Wen, Xumeng and Liu, Zihan and Zheng, Shun and Ye, Shengyu and Wu, Zhirong and Wang, Yang and Xu, Zhijian and Liang, Xiao and Li, Junjie and Miao, Ziming and others},
  journal={arXiv preprint arXiv:2506.14245},
  year={2025}
}

@article{bhaskar2025language,
  title={Language models that think, chat better},
  author={Bhaskar, Adithya and Ye, Xi and Chen, Danqi},
  journal={arXiv preprint arXiv:2509.20357},
  year={2025}
}

@article{xu2023imagereward,
  title={Imagereward: Learning and evaluating human preferences for text-to-image generation},
  author={Xu, Jiazheng and Liu, Xiao and Wu, Yuchen and Tong, Yuxuan and Li, Qinkai and Ding, Ming and Tang, Jie and Dong, Yuxiao},
  journal={Advances in Neural Information Processing Systems},
  volume={36},
  pages={15903--15935},
  year={2023}
}

@article{liu2025flow,
  title={Flow-grpo: Training flow matching models via online rl},
  author={Liu, Jie and Liu, Gongye and Liang, Jiajun and Li, Yangguang and Liu, Jiaheng and Wang, Xintao and Wan, Pengfei and Zhang, Di and Ouyang, Wanli},
  journal={arXiv preprint arXiv:2505.05470},
  year={2025}
}

@article{gong2025onereward,
  title={Onereward: Unified mask-guided image generation via multi-task human preference learning},
  author={Gong, Yuan and Wang, Xionghui and Wu, Jie and Wang, Shiyin and Wang, Yitong and Wu, Xinglong},
  journal={arXiv preprint arXiv:2508.21066},
  year={2025}
}

@inproceedings{wu2025janus,
  title={Janus: Decoupling visual encoding for unified multimodal understanding and generation},
  author={Wu, Chengyue and Chen, Xiaokang and Wu, Zhiyu and Ma, Yiyang and Liu, Xingchao and Pan, Zizheng and Liu, Wen and Xie, Zhenda and Yu, Xingkai and Ruan, Chong and others},
  booktitle={Proceedings of the Computer Vision and Pattern Recognition Conference},
  pages={12966--12977},
  year={2025}
}

@article{zhou2024transfusion,
  title={Transfusion: Predict the next token and diffuse images with one multi-modal model},
  author={Zhou, Chunting and Yu, Lili and Babu, Arun and Tirumala, Kushal and Yasunaga, Michihiro and Shamis, Leonid and Kahn, Jacob and Ma, Xuezhe and Zettlemoyer, Luke and Levy, Omer},
  journal={arXiv preprint arXiv:2408.11039},
  year={2024}
}

@article{deng2025emerging,
  title={Emerging properties in unified multimodal pretraining},
  author={Deng, Chaorui and Zhu, Deyao and Li, Kunchang and Gou, Chenhui and Li, Feng and Wang, Zeyu and Zhong, Shu and Yu, Weihao and Nie, Xiaonan and Song, Ziang and others},
  journal={arXiv preprint arXiv:2505.14683},
  year={2025}
}

@article{chen2025blip3,
  title={Blip3-o: A family of fully open unified multimodal models-architecture, training and dataset},
  author={Chen, Jiuhai and Xu, Zhiyang and Pan, Xichen and Hu, Yushi and Qin, Can and Goldstein, Tom and Huang, Lifu and Zhou, Tianyi and Xie, Saining and Savarese, Silvio and others},
  journal={arXiv preprint arXiv:2505.09568},
  year={2025}
}

@article{li2024autoregressive,
  title={Autoregressive image generation without vector quantization},
  author={Li, Tianhong and Tian, Yonglong and Li, He and Deng, Mingyang and He, Kaiming},
  journal={Advances in Neural Information Processing Systems},
  volume={37},
  pages={56424--56445},
  year={2024}
}

@article{wu2025harmonizing,
  title={Harmonizing visual representations for unified multimodal understanding and generation},
  author={Wu, Size and Zhang, Wenwei and Xu, Lumin and Jin, Sheng and Wu, Zhonghua and Tao, Qingyi and Liu, Wentao and Li, Wei and Loy, Chen Change},
  journal={arXiv preprint arXiv:2503.21979},
  year={2025}
}

@article{zhao2024monoformer,
  title={Monoformer: One transformer for both diffusion and autoregression},
  author={Zhao, Chuyang and Song, Yuxing and Wang, Wenhao and Feng, Haocheng and Ding, Errui and Sun, Yifan and Xiao, Xinyan and Wang, Jingdong},
  journal={arXiv preprint arXiv:2409.16280},
  year={2024}
}

@article{chen2024diffusion,
  title={Diffusion forcing: Next-token prediction meets full-sequence diffusion},
  author={Chen, Boyuan and Mart{\'\i} Mons{\'o}, Diego and Du, Yilun and Simchowitz, Max and Tedrake, Russ and Sitzmann, Vincent},
  journal={Advances in Neural Information Processing Systems},
  volume={37},
  pages={24081--24125},
  year={2024}
}

@inproceedings{wu2024self,
  title={Self-correcting llm-controlled diffusion models},
  author={Wu, Tsung-Han and Lian, Long and Gonzalez, Joseph E and Li, Boyi and Darrell, Trevor},
  booktitle={Proceedings of the IEEE/CVF Conference on Computer Vision and Pattern Recognition},
  pages={6327--6336},
  year={2024}
}

@article{cao2023beautifulprompt,
  title={Beautifulprompt: Towards automatic prompt engineering for text-to-image synthesis},
  author={Cao, Tingfeng and Wang, Chengyu and Liu, Bingyan and Wu, Ziheng and Zhu, Jinhui and Huang, Jun},
  journal={arXiv preprint arXiv:2311.06752},
  year={2023}
}

@inproceedings{wallace2024diffusion,
  title={Diffusion model alignment using direct preference optimization},
  author={Wallace, Bram and Dang, Meihua and Rafailov, Rafael and Zhou, Linqi and Lou, Aaron and Purushwalkam, Senthil and Ermon, Stefano and Xiong, Caiming and Joty, Shafiq and Naik, Nikhil},
  booktitle={Proceedings of the IEEE/CVF Conference on Computer Vision and Pattern Recognition},
  pages={8228--8238},
  year={2024}
}

@inproceedings{su2024ddpo,
  title={DDPO: Direct dual propensity optimization for post-click conversion rate estimation},
  author={Su, Hongzu and Meng, Lichao and Zhu, Lei and Lu, Ke and Li, Jingjing},
  booktitle={Proceedings of the 47th International ACM SIGIR Conference on Research and Development in Information Retrieval},
  pages={1179--1188},
  year={2024}
}

@article{wu2025rewarddance,
  title={Rewarddance: Reward scaling in visual generation},
  author={Wu, Jie and Gao, Yu and Ye, Zilyu and Li, Ming and Li, Liang and Guo, Hanzhong and Liu, Jie and Xue, Zeyue and Hou, Xiaoxia and Liu, Wei and others},
  journal={arXiv preprint arXiv:2509.08826},
  year={2025}
}

@article{ziegler2019fine,
  title={Fine-tuning language models from human preferences},
  author={Ziegler, Daniel M and Stiennon, Nisan and Wu, Jeffrey and Brown, Tom B and Radford, Alec and Amodei, Dario and Christiano, Paul and Irving, Geoffrey},
  journal={arXiv preprint arXiv:1909.08593},
  year={2019}
}

@article{kingma2013auto,
  title={Auto-encoding variational bayes},
  author={Kingma, Diederik P and Welling, Max},
  journal={arXiv preprint arXiv:1312.6114},
  year={2013}
}

@article{xie2025reconstruction,
  title={Reconstruction alignment improves unified multimodal models},
  author={Xie, Ji and Darrell, Trevor and Zettlemoyer, Luke and Wang, XuDong},
  journal={arXiv preprint arXiv:2509.07295},
  year={2025}
}

@article{loshchilov2017decoupled,
  title={Decoupled weight decay regularization},
  author={Loshchilov, Ilya and Hutter, Frank},
  journal={arXiv preprint arXiv:1711.05101},
  year={2017}
}

@article{li2023seed,
  title={Seed-bench: Benchmarking multimodal llms with generative comprehension},
  author={Li, Bohao and Wang, Rui and Wang, Guangzhi and Ge, Yuying and Ge, Yixiao and Shan, Ying},
  journal={arXiv preprint arXiv:2307.16125},
  year={2023}
}

@inproceedings{fu2025mme,
  title={Mme: A comprehensive evaluation benchmark for multimodal large language models},
  author={Fu, Chaoyou and Chen, Peixian and Shen, Yunhang and Qin, Yulei and Zhang, Mengdan and Lin, Xu and Yang, Jinrui and Zheng, Xiawu and Li, Ke and Sun, Xing and others},
  booktitle={The Thirty-ninth Annual Conference on Neural Information Processing Systems Datasets and Benchmarks Track},
  year={2025}
}

@article{li2023evaluating,
  title={Evaluating object hallucination in large vision-language models},
  author={Li, Yifan and Du, Yifan and Zhou, Kun and Wang, Jinpeng and Zhao, Wayne Xin and Wen, Ji-Rong},
  journal={arXiv preprint arXiv:2305.10355},
  year={2023}
}

@inproceedings{hudson2019gqa,
  title={Gqa: A new dataset for real-world visual reasoning and compositional question answering},
  author={Hudson, Drew A and Manning, Christopher D},
  booktitle={Proceedings of the IEEE/CVF conference on computer vision and pattern recognition},
  pages={6700--6709},
  year={2019}
}

@inproceedings{yue2024mmmu,
  title={Mmmu: A massive multi-discipline multimodal understanding and reasoning benchmark for expert agi},
  author={Yue, Xiang and Ni, Yuansheng and Zhang, Kai and Zheng, Tianyu and Liu, Ruoqi and Zhang, Ge and Stevens, Samuel and Jiang, Dongfu and Ren, Weiming and Sun, Yuxuan and others},
  booktitle={Proceedings of the IEEE/CVF Conference on Computer Vision and Pattern Recognition},
  pages={9556--9567},
  year={2024}
}

@article{ghosh2023geneval,
  title={Geneval: An object-focused framework for evaluating text-to-image alignment},
  author={Ghosh, Dhruba and Hajishirzi, Hannaneh and Schmidt, Ludwig},
  journal={Advances in Neural Information Processing Systems},
  volume={36},
  pages={52132--52152},
  year={2023}
}

@article{hu2024ella,
  title={Ella: Equip diffusion models with llm for enhanced semantic alignment},
  author={Hu, Xiwei and Wang, Rui and Fang, Yixiao and Fu, Bin and Cheng, Pei and Yu, Gang},
  journal={arXiv preprint arXiv:2403.05135},
  year={2024}
}

@misc{openai_gpt52,
  title        = {GPT-5.2},
  author       = {{OpenAI}},
  year         = {2025},
  howpublished = {\url{https://openai.com}},
  note         = {Large multimodal language model, accessed Jan. 2026}
}

@misc{google_gemini3,
  title        = {Gemini 3},
  author       = {{Google DeepMind}},
  year         = {2025},
  howpublished = {\url{https://gemini.google.com/}},
  note         = {Next-generation large multimodal language model from Google, accessed Jan. 2026}
}

@misc{alibaba_qwen3max,
  title        = {Qwen3 Max},
  author       = {{Alibaba}},
  year         = {2025},
  howpublished = {\url{https://qwen.ai/}}, 
  note         = {Trillion-parameter AI model (Qwen3 Max) from Alibaba Cloud, accessed Jan. 2026}
}

@article{bradley1952rank,
  title={Rank analysis of incomplete block designs: I. the method of paired comparisons},
  author={Bradley, Ralph Allan and Terry, Milton E},
  journal={Biometrika},
  volume={39},
  number={3/4},
  pages={324--345},
  year={1952},
  publisher={JSTOR}
}

@article{wang2025pref,
  title={Pref-grpo: Pairwise preference reward-based grpo for stable text-to-image reinforcement learning},
  author={Wang, Yibin and Li, Zhimin and Zang, Yuhang and Zhou, Yujie and Bu, Jiazi and Wang, Chunyu and Lu, Qinglin and Jin, Cheng and Wang, Jiaqi},
  journal={arXiv preprint arXiv:2508.20751},
  year={2025}
}
\bibliographystyle{icml2026}

\newpage
\appendix
\onecolumn
\section{Appendix}

\subsection{Dataset Construction Details: Visual Instruction Elaboration}
\label{app:dataset}

To effectively bootstrap the model's generative reasoning, we constructed a compact but high-leverage dataset for the Visual Instruction Elaboration task. The dataset is structured along two orthogonal dimensions of reasoning depth, comprising six distinct semantic categories.

\subsubsection{Data Categorization and Examples}
We categorize instructions into Simple and Hard groups. Simple Instructions focus on direct attribute mapping, while Hard Instructions necessitate deep inferential reasoning. Table \ref{tab:examples} provides specific definitions and examples for each category.

\begin{table}[h]
\centering
\caption{Detailed Categories of the Visual Instruction Elaboration Dataset. We cover 6 categories ranging from explicit synthesis to abstract reasoning.}
\resizebox{\linewidth}{!}{
\begin{tabular}{l|l|l}
\toprule
\textbf{Difficulty} & \textbf{Category} & \textbf{Description \& Example} \\
\midrule
\multirow{3}{*}{\textbf{Simple}} 
& \textbf{Material Synthesis} & Explicitly changing the texture or material of the object. \\
& & \textit{Ex: ``Transform it so it is carved entirely out of ice." } \\
\cmidrule{2-3}
& \textbf{Perspective Shifts} & Altering the camera angle or visual style without changing semantics. \\
& & \textit{Ex: ``Show it through a grainy surveillance footage." } \\
\cmidrule{2-3}
& \textbf{Semantic Edits} & Adding explicit accessories or modifying basic attributes. \\
& & \textit{Ex: ``Place a golden crown on the object's head or top." } \\
\midrule
\multirow{3}{*}{\textbf{Hard}} 
& \textbf{Attribute Reasoning} & Inferring implicit visual details from underspecified descriptions. \\
& & \textit{Ex: ``Make it look more refined and carefully made." } \\
\cmidrule{2-3}
& \textbf{Constraint Reasoning} & Resolving semantic conflicts or balancing paradoxical constraints. \\
& & \textit{Ex: ``Make the object look scary, yet undeniably cute." } \\
\cmidrule{2-3}
& \textbf{Conceptual Reasoning} & Grounding high-level visual concepts into concrete visual elements. \\
& & \textit{Ex: ``Make this image more memorable." } \\
\bottomrule
\end{tabular}
}
\label{tab:examples}
\end{table}

\subsubsection{Data Collection Pipeline (Training)}
\label{trainset}
We constructed the training set using a scalable, minimal-supervision strategy to balance the quality and efficiency of self-evolving. The pipeline consists of three steps:
\begin{enumerate}
    \item \textbf{Candidate Generation:} We generate a diverse pool of 600 candidate samples (spanning 6 categories with 10 instructions per category).
    \item \textbf{Image-Based Feasibility Filtering:} To ensure the training instructions were visually realizable given the UMM's base capability, we generated 600 candidate images from the initial pairs. We then applied a dual-filter mechanism to evaluate the alignment between the generated image, the instruction, and the prompt: (1)Automated Check: First, GPT-5 performed a semantic consistency check to reject obviously low-quality or misaligned generations. (2) Sparse Human Verification: Second, for the remaining candidates, we conducted sparse human verification, where annotators quickly validated the fine-grained alignment. Only the 300 pairs that passed both filters were retained.
    \item \textbf{Final Selection:} From the filtered pool, we retained the top 300 high-quality $(a, p_0)$ pairs for training. This process ensures the dataset is accurate enough for RLVR while minimizing human annotation costs.
\end{enumerate}

\subsubsection{Test Set Configuration}
To rigorously assess the capabilities of SEER, we constructed a dedicated test set completely separate from the training data. Crucially, unlike the training set which was filtered for visual realizability to facilitate learning, the test set was curated solely based on semantic coverage without considering the model's base capability. This ensures a fair, unbiased, and comprehensive evaluation of the model's true boundaries.
\begin{itemize}
    \item \textbf{Scale:} We explicitly selected 40 instructions.
    \item \textbf{Distribution Split:}
        \begin{itemize}
            \item In-Distribution (ID): Instructions seen during training but combined with entirely new prompts $p_0$ to test combinatorial generalization.
            \item Out-Of-Distribution (OOD): Instructions representing entirely unseen semantic concepts (e.g., new materials or abstract vibes) to test zero-shot transfer capabilities.
            \item The ratio of ID to OOD is 1:1 (20 vs. 20).
        \end{itemize}
    \item \textbf{Difficulty Split:} The ratio of Simple to Hard instructions is 2:3, emphasizing the evaluation of complex reasoning capabilities on harder tasks.
\end{itemize}

\begin{figure*}[!t]
    \centering
    \begin{minipage}{\textwidth}
        \centering
        \includegraphics[width=0.8\textwidth]{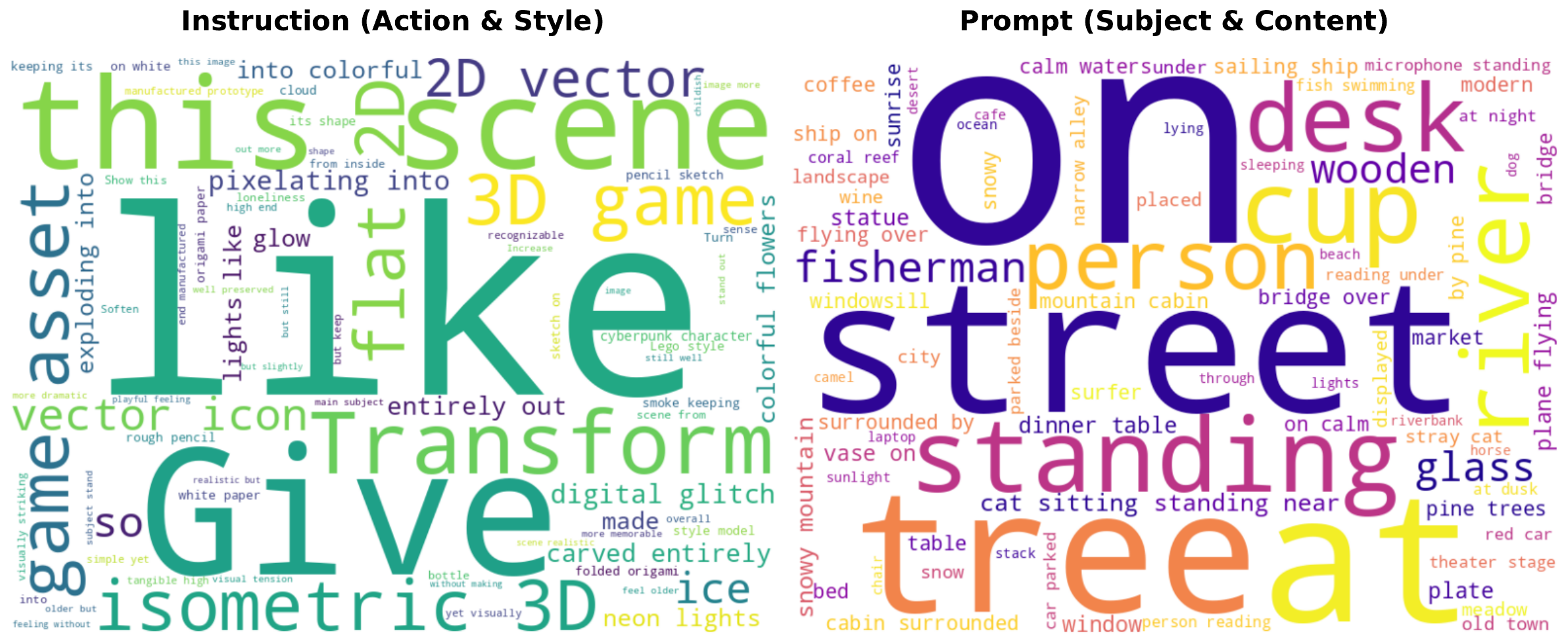}
    \end{minipage}
    
    \begin{minipage}{\textwidth}
        \centering
        \includegraphics[width=0.8\textwidth]{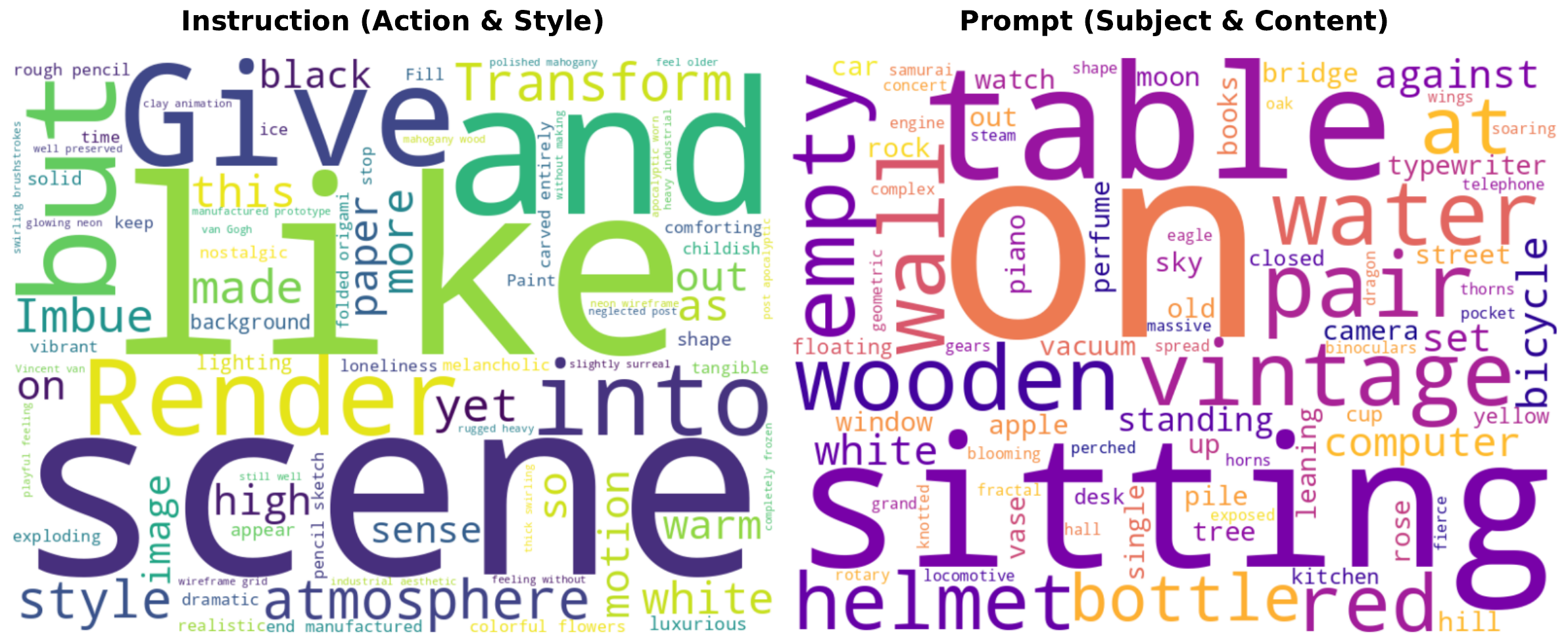}
    \end{minipage}
    
    \caption{Lexical Distribution Shift Analysis. We visualize the word clouds for Instructions (Left) and Prompts (Right) across the Training (Top) and Test (Below) sets. 
    Prompts: The significant vocabulary shift (e.g., ``street'' $\to$ ``table'') ensures rigorous testing of object generalization. 
   Instructions: While structural triggers overlap to maintain task consistency, the Test set introduces novel vocabulary (OOD concepts), challenging the model's zero-shot reasoning capabilities.}
    \label{fig:lexical_shift}
    \vspace{-0.6cm}
\end{figure*}

\subsubsection{Training Curriculum and Negative Sampling}
\label{app:curriculum}

For the RLVR (Evaluator Training) stage, we utilize the 300 training pairs $(a, p_0)$ constructed in Sec.~\ref{trainset}. We implement a two-phase curriculum to progressively activate the evaluators:

\textbf{Curriculum 1: Basic Alignment (Warm-up).} 
To teach the evaluator basic image-text alignment, we use the initial prompt $p_0$ from the training set.
\begin{itemize}
    \item \textbf{Positive Sample ($x^+_{c1}$):} The image generated directly using $p_0$.
    \item \textbf{Negative Sample ($x^-_{c1}$):} An image generated by randomly masking or adding the object or scene in $p_0$, creating a semantic mismatch.
\end{itemize}

\subsubsection{Distributional Analysis: Train vs. Test.}
\label{difference}
To strictly evaluate the model's generalization capabilities, we analyze the lexical overlap between the training and test splits, as visualized in Figure \ref{fig:lexical_shift} and Table \ref{tab:id_ood_examples}.
\begin{itemize}
    \item \textbf{Prompt Divergence (Object Generalization):} As shown in the Prompt word clouds (Right), while both sets demonstrate comprehensive semantic diversity, their vocabularies are markedly distinct. The training set covers a wide array of entities (e.g., ``street'', ``standing'', ``river''), whereas the test set spans an equally diverse but disjoint set of subjects (e.g., ``table'', ``wooden'', ``vintage''). This disjoint yet diverse distribution ensures that the model cannot simply memorize object-attribute pairs; it must learn to apply instruction-guided modifications to entirely new semantic subjects.
    \item \textbf{Instruction Expansion (Concept Generalization):} In the Instruction clouds (Left), while structural keywords (e.g., ``Transform'', ``like'') overlap to define the task format, the test set introduces a broader spectrum of semantic demands. It covers not only the seen In-Distribution (ID) types but also expands into Out-Of-Distribution (OOD) concepts (e.g., specific artistic styles or abstract atmospheres not present in training), forcing the model to perform true generative reasoning rather than pattern matching.
\end{itemize}

\textbf{Curriculum 2: Instruction Supervision (Main).} 
This phase teaches the model to verify the visual instruction $a$.
\begin{itemize}
    \item \textbf{Positive Sample ($x^+_{c2}$):} The verified ground-truth image from the training set (sampled and filtered in the data collection pipeline).
    \item \textbf{Negative Sample ($x^-_{c2}$):} We directly reuse the Positive Sample from Curriculum 1 ($x^+_{c1}$). Since this image reflects only $p_0$ without the instruction $a$, it serves as a hard negative.
\end{itemize}
The evaluator learns to prefer $x^+_{c2}$ over $x^-_{c2}$. Crucially, we maintain a clear distinction in data curation: while these training samples rely on minimal supervision for efficiency, the test set (used for Table \ref{tab:evaluator_detailed}) underwent rigorous multi-round human review to strictly ensure the Ground Truth aligns with human intuition.

\begin{table*}[h]
\centering
\caption{Distributional Shift Analysis: ID vs. OOD. We provide concrete examples demonstrating the semantic gap between In-Distribution (Seen concepts) and Out-of-Distribution (Unseen concepts) across all 6 reasoning categories.}
\resizebox{\textwidth}{!}{
\begin{tabular}{l|l|p{7cm}|p{7cm}}
\toprule
\textbf{Difficulty} & \textbf{Category} & \textbf{In-Distribution (ID) Example} \newline \textit{(Concept seen during training)} & \textbf{Out-Of-Distribution (OOD) Example} \newline \textit{(Novel concept unseen during training)} \\
\midrule
\multirow{6}{*}{\textbf{Simple}} 
& \textbf{Material Synthesis} 
& \textit{``Transform it so it is carved entirely out of \textbf{ice}."} 
& \textit{``Transform it so it is made of \textbf{polished mahogany wood}."} \\
\cmidrule{2-4}
& \textbf{Perspective Shifts} 
& \textit{``Show it through a \textbf{grainy surveillance footage}."} 
& \textit{``Render it in the style of \textbf{8-bit retro pixel art}."} \\
\cmidrule{2-4}
& \textbf{Semantic Edits} 
& \textit{``Place a \textbf{golden crown} on the object's head."} 
& \textit{``Transform the object into a \textbf{die-cut sticker with a white border}."} \\
\midrule
\multirow{6}{*}{\textbf{Hard}} 
& \textbf{Inferential Reasoning} 
& \textit{``Make it look more \textbf{refined and carefully made}."} 
& \textit{``Give it a \textbf{rugged, heavy industrial aesthetic}."} \\
\cmidrule{2-4}
& \textbf{Constraint Reasoning} 
& \textit{``Make the object look \textbf{scary, yet undeniably cute}."} 
& \textit{``Render it as \textbf{cheap plastic}, yet make it appear \textbf{precious and luxurious}."} \\
\cmidrule{2-4}
& \textbf{Conceptual Reasoning} 
& \textit{``Make this image more \textbf{memorable}."} 
& \textit{``Make this image convey a sense of \textbf{impending danger}."} \\
\bottomrule
\end{tabular}
}
\label{tab:id_ood_examples}
\end{table*}

\subsubsection{Training Curriculum and Negative Sampling}

\subsection{Human Evaluation Protocol}
\label{app:human_eval}

To calculate the Human Win Ratio, we adhered to a strict blind pairwise protocol:
\begin{itemize}
    \item \textbf{Annotators:} 6 independent human evaluators.
    \item \textbf{Procedure:} For each test case, the evaluator is presented with two images: one from SEER and one from a Baseline model. The order of images is randomized to prevent position bias.
    \item \textbf{Criterion:} ``Which image better satisfies the instruction '$a$' while preserving the subject '$p_0$'?"
    \item \textbf{Scoring:}
    \begin{itemize}
        \item If SEER is chosen: +1 point.
        \item If Baseline is chosen: 0 points.
        \item If Tie: 0.5 points.
    \end{itemize}
    \item \textbf{Final Metric:} The final Win Ratio is the average score across all annotators and all test cases.
    \item \textbf{Statistical Reliability:} To ensure the robustness of our evaluation, we conducted significance testing on the human judgment results. SEER outperforms all baselines with statistical significance. Specifically, for our dominant results (e.g., against Base), the improvement is highly significant ($p < 0.001$). Even for competitive baselines (win ratio 0.61), the win ratios remain statistically significant ($p < 0.05$), confirming that the performance gains are consistent across human annotators and not due to random variance.
\end{itemize}

\subsection{Prompt Templates}
\label{app:prompts}

We utilized structured system prompts to enforce the Chain-of-Thought (CoT) reasoning format during both training and inference.

\subsubsection{System Prompt for Reasoning}
The following system prompt guides the model to explicitize its reasoning process before generating the final reprompt.

\begin{tcolorbox}[
    colback=bg_gray,      
    colframe=border_gray,  
    boxrule=0.5pt,          
    arc=2mm,             
    left=3mm, right=3mm, top=3mm, bottom=3mm,
    fontupper=\small\ttfamily
]
A conversation between User and Assistant. \\
The user asks a question, and the Assistant solves it. \\
The assistant first thinks about the reasoning process in the mind and then provides the user with the answer. \\
The reasoning process and answer are enclosed within <think> </think> and <answer> </answer> tags, respectively, i.e., \\
<think> reasoning process here </think> \\
<answer> answer here </answer>
\end{tcolorbox}

\subsubsection{Prompt for Pairwise Evaluation (Stage 1)}
\label{reward_prompt}
During RLVR training and inference, we leverage the VLM's inherent reasoning capabilities to consolidate the three reward dimensions (defined in Eq.~\ref{eq:reward_components}) into a single holistic judgment.
As shown in the template below, the instruction ``correctly follows the instruction... while keeping the subject recognizable... and having better quality" explicitly fuses Compliance, Consistency, and Visual Quality into a unified comparison criterion.

\begin{tcolorbox}[
    colback=white!95!gray,      
    colframe=gray!50!black,  
    boxrule=0.5pt,          
    arc=2mm,             
    left=3mm, right=3mm, top=3mm, bottom=3mm,
    fontupper=\small\ttfamily
]
Here is the first image. <image> \\
Here is the second image. <image> \\
User Instruction: \{instruction\} \\
Base Prompt: \{prompt\} \\
Compare the images. The goal is to apply the Instruction to the Base Subject. \\
You MUST prefer the image that \textbf{correctly follows the instruction} (e.g., material/perspective change) while \textbf{keeping the subject recognizable} and \textbf{having better quality}. \\
Does the first image match the instruction better than the second image? \\
First output the thinking process in <think> </think> tags and then output the final answer with only 'Yes' or 'No' in <answer> </answer> tags.
\end{tcolorbox}

\subsubsection{Prompt for Endogenous Reprompting (Stage 2)}
During RLMT training and inference, the generator uses the following template to generate reprompts:

\begin{tcolorbox}[
    colback=bg_gray,      
    colframe=border_gray,  
    boxrule=0.5pt,          
    arc=2mm,             
    left=3mm, right=3mm, top=3mm, bottom=3mm,
    fontupper=\small\ttfamily
]
Rewrite the base prompt to follow the instruction while strictly preserving the identity of the core subject. \\
User Instruction: instruction\\
Base Prompt: prompt\\
1. Identify Subject: Keep the main object in the base prompt as the central focus. \\
2. Translate to Visuals: Convert the instruction into concrete visual descriptors (e.g., specific style, lighting, texture, or composition details). \\
3. Rewrite: Write the final prompt integrating these new details. \\
First output the reasoning in <think> </think> tags, then output the final image generation prompt in <answer> </answer> tags.
\end{tcolorbox}

\subsection{Qualitative Visualizations}
\label{app:viz}

\subsubsection{Consistency of Reprompting (Ablation)}
\begin{figure*}[t]
    \centering
    \includegraphics[width=0.9\textwidth]{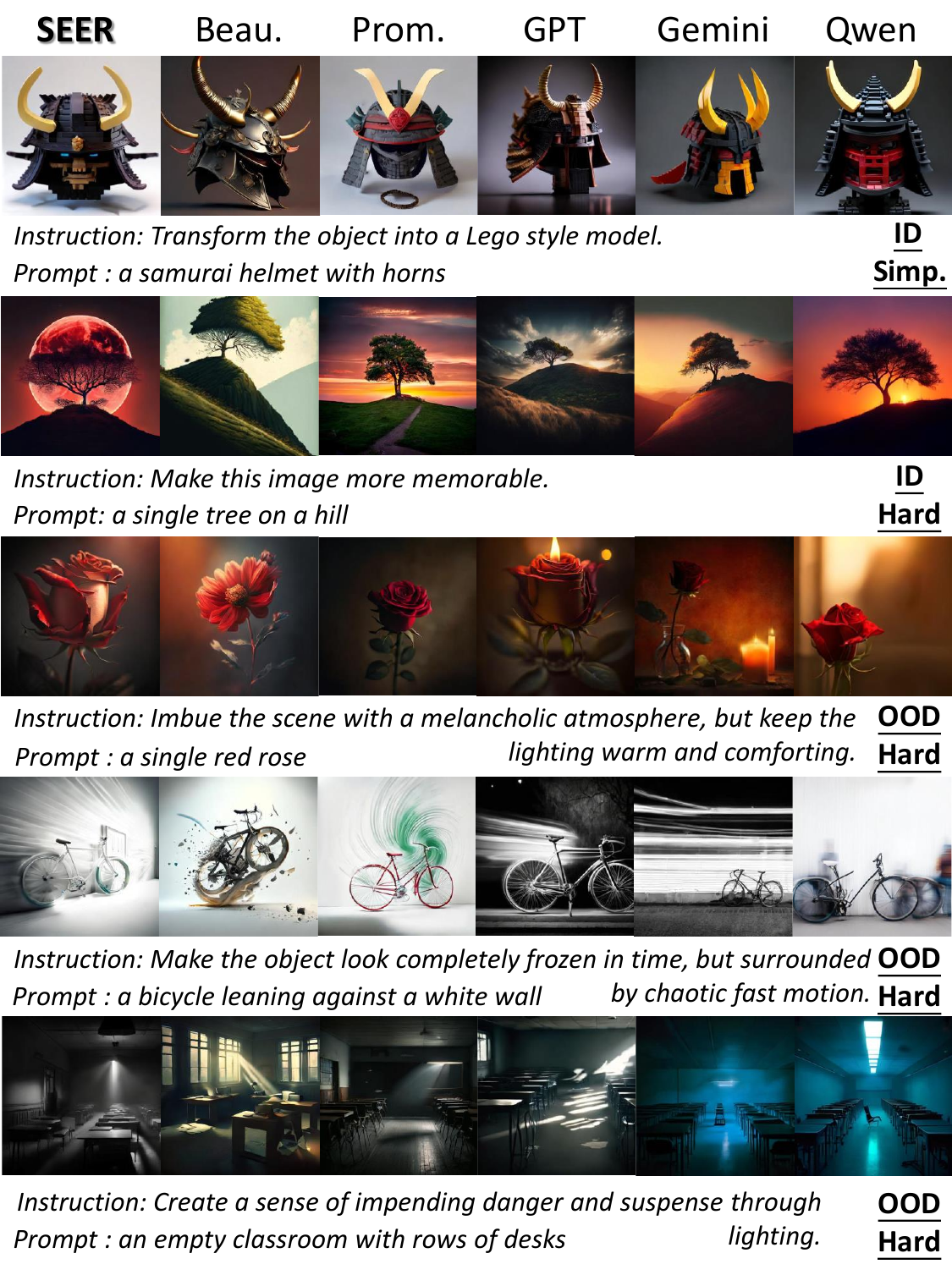}
    \caption{
    Qualitative Ablation: Impact of Reprompter Source.
    We verify model-specific alignment by feeding different reprompts into the fixed SEER-Gen. 
    While external reprompters (Middle/Right) often suffer from Instruction Neglect, rendering the prompt object literally but failing to trigger the requested instruction shift (e.g., Lego, Fast Motion). SEER (Left) successfully concretizes visual instructions into effective visual descriptions.
    }
    \label{fig:ablation_reprompt}
\end{figure*}

In Figure \ref{fig:ablation_reprompt}, we visualize the generation outcomes when feeding reprompts from different sources (e.g., GPT-5.2, PromptEnhancer, vs. SEER) into the fixed SEER-Gen. 
This qualitative ablation demonstrates that while external reprompts may be linguistically rich, they often cause the generator to miss attributes due to representation mismatch. In contrast, SEER's self-aligned reprompts effectively trigger the generator's priors, confirming the necessity of model-specific alignment.

\subsubsection{Comparison with SOTA UMMs}
\begin{figure*}[t]
    \centering
    \includegraphics[width=0.9\textwidth]{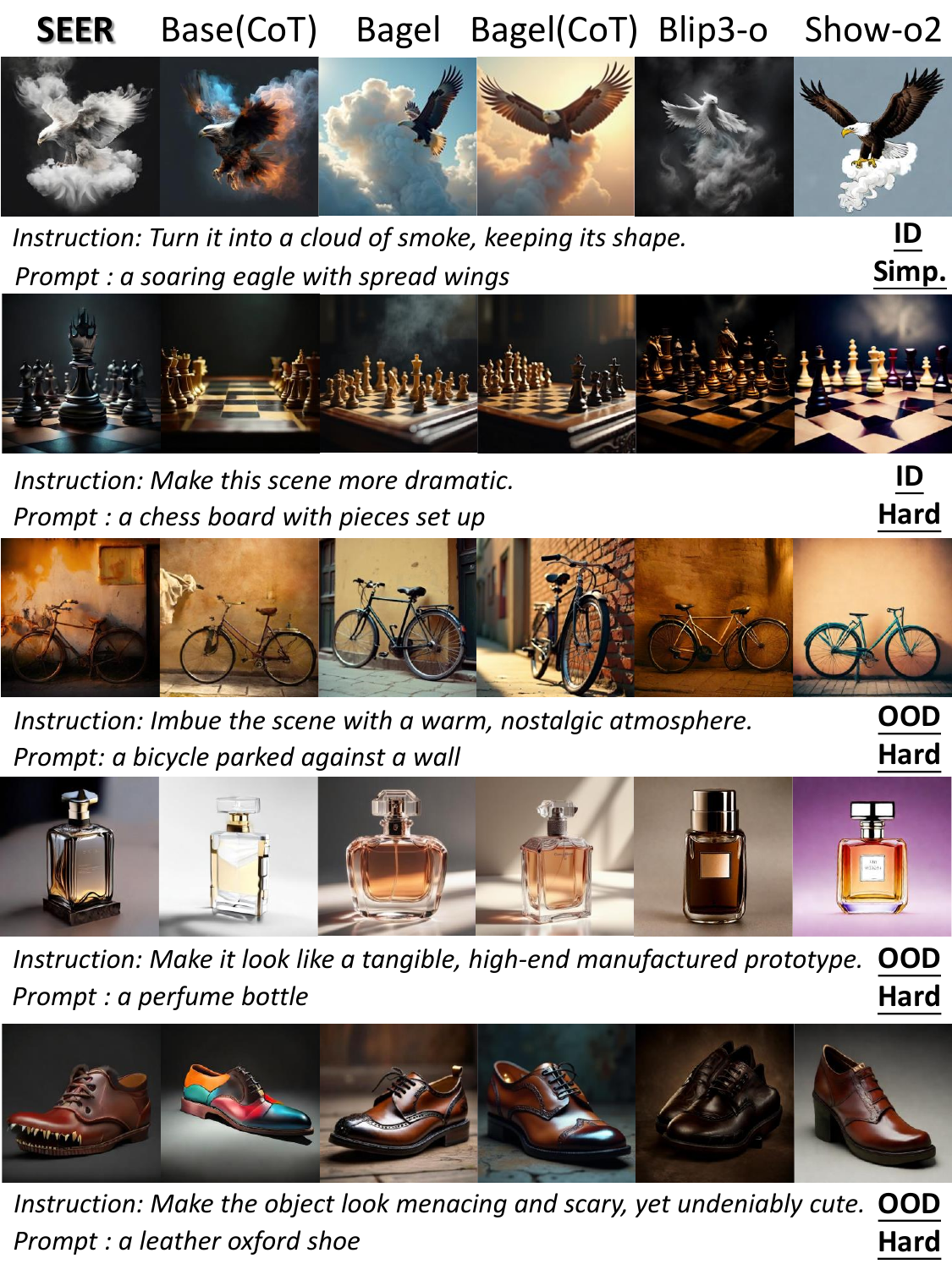}
    \caption{
    Visual Benchmarking against SOTA UMMs.
    Baselines (Middle/Right) frequently suffer from literal interpretation, ignoring visual instructions (e.g., drawing a normal eagle instead of ``smoke'', or a normal shoe instead of ``scary and cute''). 
    SEER (Left) demonstrates superior generative reasoning, successfully fusing the user intent with the subject.
    }
    \label{fig:sota_compare}
\end{figure*}

Figure \ref{fig:sota_compare} presents a gallery of comparisons between SEER and baselines (Bagel, Show-o2) on visual instructions. By leveraging generative reasoning, SEER faithfully executes the user's intent across diverse scenarios, whereas baselines often suffer from semantic collapse, attribute neglect, or literal interpretation failures.

\subsubsection{Fine-grained Analysis of the Cognitive Gap}
\label{app:failure_analysis}

To precisely characterize the cognitive gap, we categorize the typical failure modes of UMMs into three distinct types, as illustrated in Figure \ref{fig:failure_visuals}. These failures represent the specific behaviors our endogenous reward model aims to penalize, driving the policy towards more effective reasoning.

\begin{itemize}
    \item \textbf{Type I: Naive Concatenation (Laziness).} The model simply appends the instruction text or its synonyms to the prompt without semantic integration (e.g., Prompt: ``A cat", Output: ``A cat, make it cute"). This reflects a lack of active planning, resulting in images that fail to visualize the intended change.
    
    \item \textbf{Type II: Semantic Misinterpretation (Hallucination).} The model misunderstands the instruction's intent, translating it into unrelated or conflicting visual features (e.g., interpreting ``frozen in time" as ``covered in ice" instead of ``motionless"). This leads to visual hallucinations that diverge from the user's goal.
    
    \item \textbf{Type III: Instruction Neglect (Omission).} The model correctly identifies the subject ($p_0$) but completely ignores the visual modifier ($a$). The generated image depicts the base object perfectly but lacks the requested style or attribute. This is the most common manifestation of the gap in ``Hard" tasks.
\end{itemize}

\textbf{Optimization Mechanism.} 
During the RLMT stage, the policy explores various reasoning paths. Some generated CoT traces may still lead to these failure modes. 
Since the internal evaluator assigns low rewards to the images resulting from these ineffective CoTs (often scoring them lower than the naive baseline), the optimization process naturally suppresses the probability of such reasoning traces. 
Simultaneously, it reinforces the effective CoT paths that successfully bridge the gap, thereby distilling the model's reasoning capability to produce high-fidelity generation.

\subsection{Limitations and Future Work}
\label{sec:conclusion}
SEER is currently bounded by the model's latent manifold; it effectively activates dormant capabilities but cannot synthesize concepts entirely absent from pre-training.
Future directions include exploring iterative multi-turn reprompting for complex constraints, and extending the framework to jointly optimize the visual generator, enabling iterative self-evolution to progressively enhance both instruction alignment and fundamental generation quality. Furthermore, we can verify the universality of SEER by applying it to diverse UMM architectures and larger scales.

\clearpage
\raggedbottom
\onecolumn

\noindent
\includegraphics[width=1\textwidth]{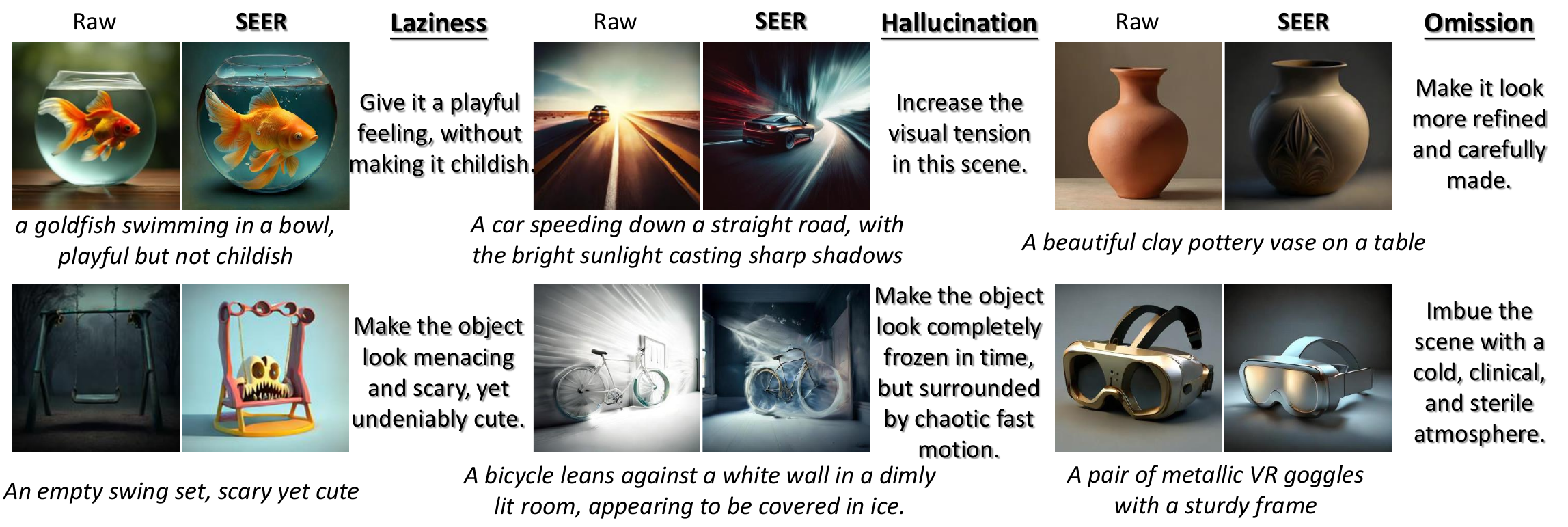}

\captionof{figure}{Visualization of Cognitive Gap Failure Modes. We compare the Raw baseline with SEER across the three failure types: (I) Laziness (naive concatenation), (II) Hallucination (semantic misinterpretation), and (III) Omission (instruction neglect). The text below each image pair represents the generated reprompts of Raw baseline. While Raw reprompts consistently fail to execute complex instructions, SEER achieves precise visual-semantic alignment through endogenous reasoning.}
\label{fig:failure_visuals}

\twocolumn


\end{document}